\documentclass[10pt,twocolumn,letterpaper]{article}
\usepackage[left]{lineno}
\usepackage[pagenumbers]{cvpr} 
\usepackage{times}
\usepackage{epsfig}
\usepackage{graphicx}
\usepackage{amsmath}
\usepackage{amssymb}

\usepackage[svgnames]{xcolor}
\usepackage{algorithm}
\usepackage{algorithmic}
\usepackage{multirow}
\usepackage{amsthm}
\usepackage{color}
\usepackage{colortbl}
\usepackage{ctable}
\usepackage{pifont}
\usepackage{ragged2e}
\usepackage{booktabs}
\usepackage{enumitem}
\usepackage{xspace}

\newcommand{\cmark}{\ding{51}}%
\newcommand{\xmark}{\ding{55}}%

\usepackage[pagebackref=true,breaklinks=true,letterpaper=true,colorlinks,bookmarks=false]{hyperref}

\newcommand{\mtl}{{multi-target knowledge distillation}\xspace}
\newcommand{\Mtl}{{Multi-target knowledge distillation}\xspace}
\newcommand{\MTL}{{MT-KD}\xspace}
\newcommand{\mta}{{unseen target knowledge distillation}\xspace}
\newcommand{\Mta}{{Unseen target knowledge distillation}\xspace}
\newcommand{\MTA}{{UT-KD}\xspace}
\newcommand{\mts}{{multi-target style transfer network}\xspace}
\newcommand{\Mts}{{Multi-target style transfer network}\xspace}
\newcommand{\MTS}{{MT-STN}\xspace}

\newcommand{\ce}{{\textsc{ce}}}
\newcommand{\con}{{\textrm{con}}}
\newcommand{\out}{{\textrm{out}}}
\newcommand{\rec}{{\textrm{rec}}}
\newcommand{\adv}{{\textrm{adv}}}
\newcommand{\cin}{{\operatorname{CIN}}}
\newcommand{\fro}{{\textrm{fro}}}

\newcommand{\expect}{\mathbb{E}}

\newcommand{\real}{\mathbb{R}}
\newcommand{\cL}{\mathcal{L}}
\newcommand{\tran}{^\top}
\newcommand{\cA}{\mathcal{A}}
\newcommand{\cX}{\mathcal{X}}
\newcommand{\cV}{\mathcal{V}}
\newcommand{\cD}{\mathcal{D}}
\newcommand{\norm}[1]{\left\|{#1}\right\|}
\newcommand{\id}{\operatorname{id}}

\newcommand{\Th}[1]{\textsc{#1}}
\newcommand{\mr}[2]{\multirow{#1}{*}{#2}}
\newcommand{\mc}[2]{\multicolumn{#1}{c}{#2}}

\newcommand{\eq}[1]{(\ref{eq:#1})}
\newcommand{\fig}[2][1]{\includegraphics[width=#1\linewidth]{fig/#2}}

\begin{document}

\title{Multi-Target Unsupervised Domain Adaptation for Semantic Segmentation without External Data}

\author{Yonghao Xu$^{1,2}$, Pedram Ghamisi$^{1,3}$, Yannis Avrithis$^{1}$\\
$^{1}$Institute of Advanced Research in Artificial Intelligence (IARAI), Austria\\
$^{2}$Computer Vision Laboratory, Link\"{o}ping University, Sweden\\
$^{3}$Helmholtz Institute Freiberg for Resource Technology, HZDR, Germany\\
{\tt\small yonghao.xu@liu.se, p.ghamisi@hzdr.de, yannis@avrithis.net
}
}
\maketitle

%%%%%%%%% ABSTRACT
\begin{abstract}
Multi-target unsupervised domain adaptation (UDA) aims to learn a unified model to address the domain shift between multiple target domains. Due to the difficulty of obtaining annotations for dense predictions, it has recently been introduced into cross-domain semantic segmentation. However, most existing solutions require labeled data from the source domain and unlabeled data from multiple target domains concurrently during training. Collectively, we refer to this data as ``external''. When faced with new unlabeled data from an unseen target domain, these solutions either do not generalize well or require retraining from scratch on all data. %This significantly increases the cost of applying them to new domains or renders them unsuitable when the original external data is inaccessible. 
To address these challenges, we introduce a new strategy called ``multi-target UDA without external data'' for semantic segmentation. Specifically, the segmentation model is initially trained on the external data. Then, it is adapted to a new unseen target domain without accessing any external data. This approach is thus more scalable than existing solutions and remains applicable when external data is inaccessible. We demonstrate this strategy using a simple method that incorporates self-distillation and adversarial learning, where knowledge acquired from the external data is preserved during adaptation through ``one-way'' adversarial learning. Extensive experiments in several synthetic-to-real and real-to-real adaptation settings on four benchmark urban driving datasets show that our method significantly outperforms current state-of-the-art solutions, even in the absence of external data. Our source code is available online (https://github.com/YonghaoXu/UT-KD).
\end{abstract}

\section{Introduction}
\label{sec:intro}

%------------------------------------------------------------------------------
\begin{figure}
\centering
\fig[1]{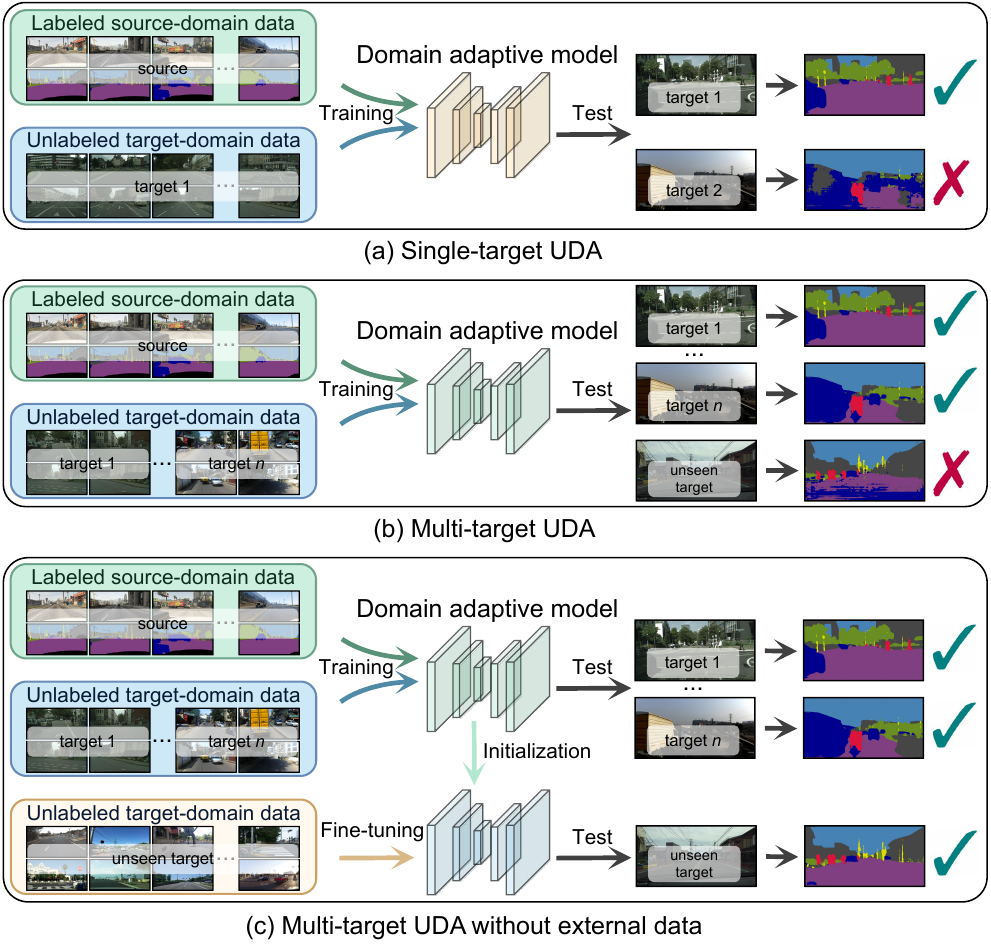}
\caption{Different strategies in cross-domain semantic segmentation. (a) \emph{Single-target unsupervised domain adaptation} (UDA): the segmentation model cannot generalize well to unseen domains. (b) \emph{Multi-target UDA}: target domains are still predetermined at training and the model needs to be retrained from scratch on all data when a new unseen target domain is given, or else it will suffer from the same problem. (c) Our new strategy, \emph{multi-target UDA without external data}: the pre-trained model is quickly adapted to a new unseen target domain without accessing any external data from the original source or target domains.}
\label{fig:demo}
\end{figure}
%------------------------------------------------------------------------------

Among many other computer vision tasks, progress in deep learning has significantly advanced semantic segmentation~\cite{deeplab}. Nevertheless, the particular difficulty of semantic segmentation, bing a dense prediction task, is that training a deep learning-based model usually requires a large amount of pixel-level annotations, which are laborious and time-consuming to collect. To address this challenge and mitigate the insufficiency of labeled data, \emph{unsupervised domain adaptation} (UDA) algorithms have been recently developed for \emph{cross-domain semantic segmentation} \cite{huang2022category,lee2022wildnet}. The latter aims to learn a domain-adaptive segmentation model by training on \emph{labeled} source-domain data and \emph{unlabeled} target-domain data \cite{lian2019constructing}.

So far, most of the existing UDA methods are designed for a \emph{single target} domain \cite{zhang2017curriculum}. The main limitation of such methods is that the segmentation model may perform well in the target domain they are trained on, but can hardly generalize well to other unseen domains~\cite{isobe2021multi}. Consequently, they are unable to perform effectively, for example, on urban driving data from different cities with diverse visual styles and imaging environments, as demonstrated in \autoref{fig:demo}(a). A natural idea to address this problem is to extend to \emph{multi-target} UDA \cite{saporta2021multi,lee2022adas}. This involves adapting a labeled source domain to multiple unlabeled target domains, as illustrated in \autoref{fig:demo}(b). However, since the target domains are predetermined during training, the model either does not generalize well to unseen domains or requires retraining from scratch on all data to do so. This significantly increases the cost of applying these approaches to new domains and renders them unsuitable when the original external data is inaccessible.

To address these challenges, we introduce a new strategy called \emph{multi-target UDA without external data} for semantic segmentation. In particular, the segmentation model is first trained on labeled source-domain and unlabeled target-domain data from multiple targets. This data is collectively referred to as \emph{external}. Then, the pre-trained segmentation model is adapted to a new unseen target domain without accessing any external data, as shown in \autoref{fig:demo}(c). This strategy leverages the knowledge of the pre-trained model to eliminate the dependency on external data. Therefore, it is more scalable compared to existing multi-target UDA approaches and remains applicable even when external data is inaccessible.

Our contributions are summarized as follows:
\begin{enumerate}[itemsep=2pt, parsep=0pt, topsep=3pt]
	\item We introduce a new multi-target UDA strategy for semantic segmentation, where the segmentation model is adapted quickly to an unseen domain using unlabeled data of this domain alone, \emph{without external data}.
	\item To exhibit this strategy, we design a simple method called \emph{\mtl} (\MTL), which uses self-distillation and adversarial learning and achieves new state-of-the-art performance on multi-target UDA for semantic segmentation.
	\item As a second step, we modify \MTL by removing access to labeled data and supervision. This new method, called \emph{\mta} (\MTA), directly adapts a pre-trained \MTL model to a new unseen target domain by a novel \emph{one-way} adversarial learning strategy, without external data. To the best of our knowledge, we are the first to do so.
	\item To further boost performance, we perform visual style transfer across multiple domains. We parameterize the style of each domain by a single vector, thus decoupling it from the style transfer process itself. The latter is performed by a \emph{\mts} (\MTS), which is shared across all domains.

\end{enumerate}

\section{Related work}
\label{sec:related}

\subsection{Single-target unsupervised domain adaptation}
Early research for cross-domain semantic segmentation primarily focuses on the adaptation of source-domain knowledge to a specific target domain. The prevailing approach commonly employed involves acquiring domain-invariant representations through the use of \emph{adversarial learning}. This adaptation process can occur within various spaces, such as the intermediate feature space \cite{luo2019significance}, the output feature space \cite{tsai2018learning}, or within the realm of fine-grained categorical features \cite{luo2019taking}. Given that the primary discrepancy among different domains lies in their visual appearances, an alternative strategy involves the application of \emph{visual style transfer} to directly mitigate the domain disparity \cite{zhang2018fully}. Nonetheless, these methodologies, while effective within their intended single target domain, tend to exhibit limited generalization capabilities across unseen domains \cite{isobe2021multi}.

\subsection{Multi-target unsupervised domain adaptation}
To address the limitation of single-target UDA, Isobe \etal~\cite{isobe2021multi} propose the first \emph{multi-target UDA} approach for semantic segmentation. In particular, they first train an expert model for each source-target pair and then conduct collaborative learning with each expert model to achieve adaptation between different target domains. Saporta \etal~\cite{saporta2021multi} further extend adversarial learning into the multi-target UDA setting, where one discriminator for each target domain aims to discriminate that domain from all other target domains. To achieve more efficient multi-target UDA, Lee \etal~\cite{lee2022adas} directly adapt a single model to multiple target domains without training multiple domain-specific expert models. However, since the aforementioned multi-target UDA methods are trained on predetermined multiple target domains, the entire model still needs to be retrained from scratch on all data when a new unseen target domain is given; otherwise, it will suffer from the same limitation of single-target UDA: it will not generalize well. This makes it difficult to apply these approaches to unseen domains.

%------------------------------------------------------------------------------
\begin{table*}
\centering
\caption{Characteristics of diverse problem settings in cross-domain semantic segmentation. $X_s$: source data; $\cX_t = \{X_{t_n}\}_{n=1}^N$: target data; $X_u$: ``unseen'' data used as target at inference, possibly after fine-tuning. \Th{Ext}: using external data ($X_s$ or $\cX_t$) while training on $X_u$, either at pre-training or fine-tuning. Single-target and multi-target UDA have to ``see'' $X_u$ at pre-training. Source-free DA and domain generalization do not use any domain other than $X_s$ and $X_u$. Example: G: GTA5; C: CityScapes; I: IDD.}
\vspace{-.5em}
\resizebox{\linewidth}{!}{
\begin{tabular}{lcccccc} \toprule
\mr{2}{\Th{Setting}} & \mc{2}{\Th{Pre-Training}} & \mr{2}{\Th{Fine-tune}} & \mr{2}{\Th{Ext}} & \mr{2}{\Th{Knowledge flow}} & \mr{2}{\Th{Example}} \\ \cmidrule{2-3}
& \Th{Source} & \Th{Target} &&&& \\ \midrule
Single-target UDA& $X_s$ & $X_{u}$ & -- & \cmark&$X_s\to X_{u}$&G$\to$I\\
Multi-target UDA& $X_s$ & $\cX_t \cup \{X_u\}$ & -- & \cmark & $X_s\to \cX_t \cup \{X_u\}$ & G$\to$\{C, I\} \\
Source-free DA& $X_s$ & -- & $X_u$ & \xmark &$X_s\to X_u$&G$\to$I\\
Domain generalization & $X_s$ & -- & -- & \xmark&$X_s\to X_u$&G$\to$I  \\ \midrule \rowcolor{LightCyan}
Multi-target UDA w/o external data (ours) & $X_s$ &$\cX_t$ & $X_u$&\xmark& ($X_s\to \cX_t)\to X_u$ & (G$\to$C)$\to$I \\ 
\bottomrule
\end{tabular}}
\label{tab:setting}
\end{table*}
%------------------------------------------------------------------------------

\subsection{Source-free domain adaptation}
While UDA approaches typically necessitate access to labeled data from the source domain and unlabeled data from the target domain, the practical application of these approaches might be hindered by potential privacy issues that could undermine the availability of source data. In such scenarios, an alternative strategy is to directly transfer the knowledge from a segmentation model pre-trained on the source domain to the target domain. This setting is known as source-free domain adaptation \cite{Teja2021uncertainty}. Liu \etal~\cite{liu2021source} propose the first source-free domain adaptation approach for semantic segmentation. Specifically, their approach involves self-supervised learning on the target domain with both pixel- and patch-level optimization objectives. Huang \etal~\cite{huang2021model} further propose a historical contrastive learning framework using a historical source hypothesis to compensate for absent source data. Kundu \etal~\cite{kundu2021generalize} use a multi-head generalization framework with self-training. All of these methods solely draw knowledge from a single source domain, as they operate under the assumption that only the pre-trained segmentation model from the source domain is at their disposal. Consequently, the transfer of knowledge from both the source domain and other known target domains remains a non-trivial challenge.

\subsection{Domain generalization}
In contrast to DA, domain generalization aims to enhance a segmentation model's ability to perform effectively in new, unseen domains. This improvement is achieved without utilizing data from the target domain during training; instead, one or more source domains are employed. Common strategies employed for domain generalization include learning domain-agnostic feature representations \cite{choi2021robustnet,lee2022wildnet} and style augmentation \cite{zhong2022adversarial}. Despite the simplicity of these methods, their performance is relatively restricted as they neglect to incorporate any data from the target domain during the training phase.

\section{Problem formulation}
\label{sec:prob}
Formally, let $X_s$ and $\cX_t=\{X_{t_n}\}_{n=1}^N$ denote the labeled \emph{source-domain} data and the unlabeled \emph{target-domain} data, respectively, where $N$ is the number of target domains. This data is collectively called \emph{external}. The source domain data $X_s$ contains pairs of the form $(x,y)$, where $x \in [0,1]^K$ is an input gray-scale image and $y \in \real^{K \times C}$ is the corresponding dense one-hot encoded class label; $K$ is the number of pixels and $C$ is the number of classes in the segmentation task. The target domain data $\cX_t$ contains only unlabeled images $x \in [0,1]^K$. Let $X_u$ denote the new \emph{unseen target-domain} data used at inference, consisting of unlabeled images $x \in [0,1]^K$.

Table~\ref{tab:setting} summarizes the characteristics of different problem settings in cross-domain semantic segmentation. Single-target UDA, multi-target UDA, and domain generalization are one-stage methods, while source-free DA and the proposed multi-target UDA without external data include a second stage of fine-tuning on $X_u$ after pre-training. We use \Th{Ext} to refer to using external data ($X_s$ or $\cX_t$) while training on $X_u$ either at pre-training or fine-tuning. The detailed formulation of each setting follows.

\emph{Single-target UDA} aims to learn a domain adaptive segmentation model $F$ using the labeled source-domain data $X_s$ and the unlabeled target-domain data $X_t$. Since we aim to conduct inference on the unseen target-domain data $X_u$ in this study, the target-domain data $X_t$ will become $X_u$ for single-target UDA methods. The knowledge flow is thereby from the source domain $X_s$ to the ``unseen'' target domain $X_u$, which has to be available at pre-training with $X_s$.

\emph{Multi-target UDA} is trained with $X_s$ and multiple target-domain data $\cX_t=\{X_{t_n}\}_{n=1}^N$. To adapt to the new unseen target domain, $X_u$ will be regarded as the $(N+1)$-th target domain for multi-target UDA methods and the complete target domain data used at pre-training will become $\cX_t \cup \{X_u\}$. Accordingly, the knowledge flow is from the source domain to multiple target domains: $X_s\to \cX_t \cup \{X_u\}$. Again, $X_u$ has to be available at pre-training with $X_s$ and $\cX_t$.

\emph{Source-free DA} involves two stages. In the first stage, a segmentation model $F$ is pre-trained on the source domain $X_s$. In the second stage, $F$ is fine-tuned on the unseen target domain $X_u$. Thus, the knowledge flow is solely from the source domain to the unseen target domain: $X_s\to X_u$. The transfer of knowledge from other known target domains (\ie, $\cX_t$) is not feasible in this case.

\emph{Domain generalization} aims to enhance a segmentation model's generalization ability to other unseen domains by training with $X_s$ alone without seeing any target-domain data. The knowledge flow is also solely from the source domain to the unseen target domain: $X_s\to X_u$, in this case without even adapting to $X_u$.

The proposed new strategy, \emph{multi-target UDA without external data}, involves two stages. In the first pre-training stage, it learns a domain adaptive segmentation model $F$ on $X_s$ and $\cX_t$. After pre-training, it is expected that the obtained model $F$ inherits knowledge from both $X_s$ and $\cX_t$. In the second stage, only the pre-trained model $F$ and $X_u$ are available. The aim is to to distill the knowledge in $F$ and adapt it to $X_u$ without accessing any external data from $X_s$ and $\cX_t$. Thus, the knowledge flow is first from the source domain to multiple known target domains, then to the new unseen target domain: ($X_s\to \cX_t)\to X_u$.

In this sense, the new strategy is similar to multi-target UDA in using multiple targets $\cX_t$, thus acquiring as much knowledge as is available, and to source-free DA in fine-tuning on $X_u$ without access to external data, thus quickly adapting to new unseen domains.

\section{Methodology}
\label{sec:method}

Here, we introduce our methodology and the particular implementation of our new strategy, \emph{multi-target UDA without external data}. We first describe our \emph{\mtl} (\MTL) method in detail, which uses self-distillation and adversarial learning for multi-target UDA. We then simplify it to derive our new \emph{\mta} (\MTA) method, which quickly adapts a pre-trained \MTL model to an unseen target domain, without accessing any external data from the original source or any other target domain. Finally, we introduce a new \emph{\mts} (\MTS) to achieve visual style transfer across multiple domains, which can serve as an add-on component for style augmentation.

%------------------------------------------------------------------------------
\begin{figure}
\centering
\fig[1]{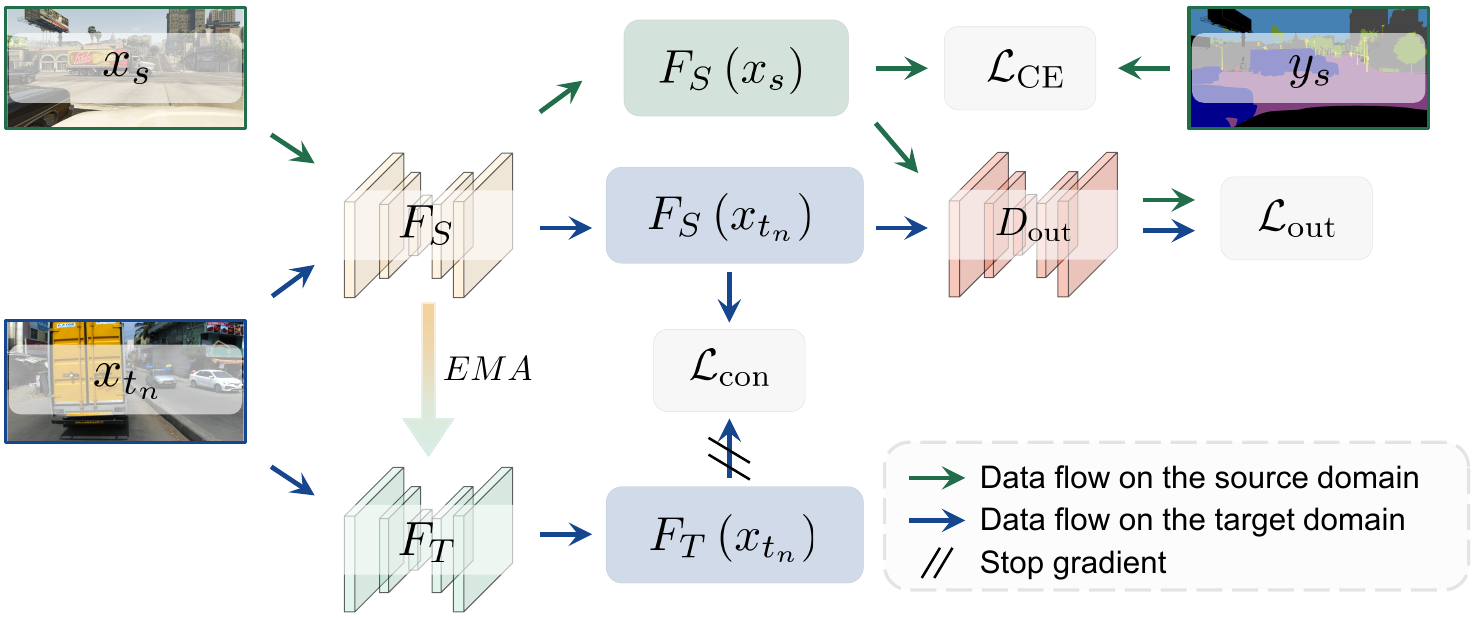}
\caption{Illustration of our \emph{\mtl} (\MTL). Given a set of labeled images $X_s$ from the source domain and unlabeled images $\cX_t=\{X_{t_n}\}_{n=1}^N$ from multiple target domains, the student network $F_S$ is trained by cross-entropy $\cL_\ce$ on the source domain, consistency loss $\cL_\con$ on the target domains and adversarial loss $\cL_\out$ in the output space. The teacher network $F_T$ is obtained by the exponential moving average (EMA) of $F_S$ parameters. Only one target domain is shown for brevity.}
\label{fig:mtkd}
\end{figure}
%------------------------------------------------------------------------------

%------------------------------------------------------------------------------
\subsection{\Mtl}
\label{sec:mtl}

As shown in \autoref{fig:mtkd}, the key idea of \emph{\mtl} (\MTL) is to conduct self-distillation and adversarial learning across multiple target domains, so that the knowledge from the labeled source domain is distilled and adapted to multiple target domains.

Formally, we aim to learn a \emph{student network} $F_S$, using a \emph{teacher network} $F_T$ of the same architecture, whose parameters $\phi_T^i$ at iteration $i$ are obtained by exponential moving average (EMA)~\cite{tarvainen2017mean} on the parameters of the student $\phi_S^i$:
\begin{linenomath}
\begin{align}
	\phi_T^i = \alpha \phi_T^{i-1} + (1-\alpha) \phi_S^i,
\label{eq:ema}
\end{align}
\end{linenomath}
where $\alpha$ is a decay parameter. Both networks are functions of the form $F: \real^{K \times 3} \to \real^{K \times C}$, which map an input image $x \in [0,1]^K$ to a predicted segmentation map $F(x) \in \real^{K \times C}$. The vector $F(x)^{(k)} \in \real^C$ is a distribution of predicted class probabilities at pixel $k$ and $F(x)^{(k,c)} \in \real$ is the predicted probability for class $c$ at pixel $k$.

On the labeled source domain data $X_s$, we define the supervised dense cross-entropy loss
\begin{linenomath}
\begin{align}
	\cL_\ce(X_s,F_S) &= \expect_{(x,y) \sim X_s} \ell_\ce(y, F_S(x)) \label{eq:ce-1} \\
	\ell_\ce(y,q) &=
		-\frac{1}{K} \sum_{k=1}^K (y^{(k)})\tran \log q^{(k)}.        \label{eq:ce-2}
\end{align}
\end{linenomath}
To distill knowledge from the labeled source domain to multiple unlabeled target domains, we apply the consistency regularization to the student predictions on unlabeled examples from multiple target domains by minimizing their mean squared error (MSE) from the teacher predictions:
\begin{linenomath}
\begin{align}
\cL_\con(\cX_t,F_S) &=
	\sum_{n=1}^{N} \expect_{x \sim X_{t_n}} \ell_\con(\cA(x), F_S) \label{eq:con-1} \\
\ell_\con(x, F_S) &=
	\frac{1}{K} \sum_{k=1}^K \norm{F_S(x)^{(k)} - F_T(x)^{(k)}}^2, \label{eq:con-2}
\end{align}
\end{linenomath}
where $\cA$ is an input transformation for data augmentation. In practice, we adopt CutMix~\cite{french2019semi} along with the proposed style transfer network \MTS. See \autoref{sec:ablation} for more details on the effect of different choices.

To encourage the $F_S$ to yield domain-invariant segmentation maps, we further introduce a discriminator $D_\out$ with a DCGAN architecture~\cite{radford2015unsupervised} to perform adversarial learning in the output space across the source and multiple target domains. In particular, the adversarial loss is defined as
\begin{multline}
	\cL_\out(X_s,\cX_t,F_S,D_\out) =\\
	\cL^+(X_s,F_S,D_\out) + \sum_{n=1}^N \cL^-(X_{t_n},F_S,D_\out),
\label{eq:out}
\end{multline}
where the two terms
\begin{linenomath}
\begin{align}
	\cL^+(X,F,D) &= \expect_{x\sim X} \log (1-D(F(x))) \label{eq:adv-pos} \\
	\cL^-(X,F,D) &= \expect_{x\sim X} \log D(F(x))     \label{eq:adv-neg}
\end{align}
\end{linenomath}
respectively represent the loss for original examples in each domain that are treated as positive for the discriminator of that domain, and the loss for the examples from other domains that are treated as negative accordingly.

Similar to previous adversarial learning work \cite{tsai2018learning}, we optimize \eq{out} through a min-max criterion, where $F_S$ aims to fool $D_\out$ by maximizing the probability of the target-domain predictions (segmentation maps) being classified as source-domain, while $D_\out$ aims to discriminate a source domain prediction from predictions of all target domains. The complete objective function is thus
\begin{linenomath}
\begin{equation}
	\min_{F_S} \max_{D_\out} \cL_\ce + \lambda_\con \cL_\con + \lambda_\out \cL_\out,
\label{eq:mtkd}
\end{equation}
\end{linenomath}
with factors $\lambda_\con$ and $\lambda_\out$ controlling the balance between the two terms.

%------------------------------------------------------------------------------
\begin{figure}
\centering
\fig[1]{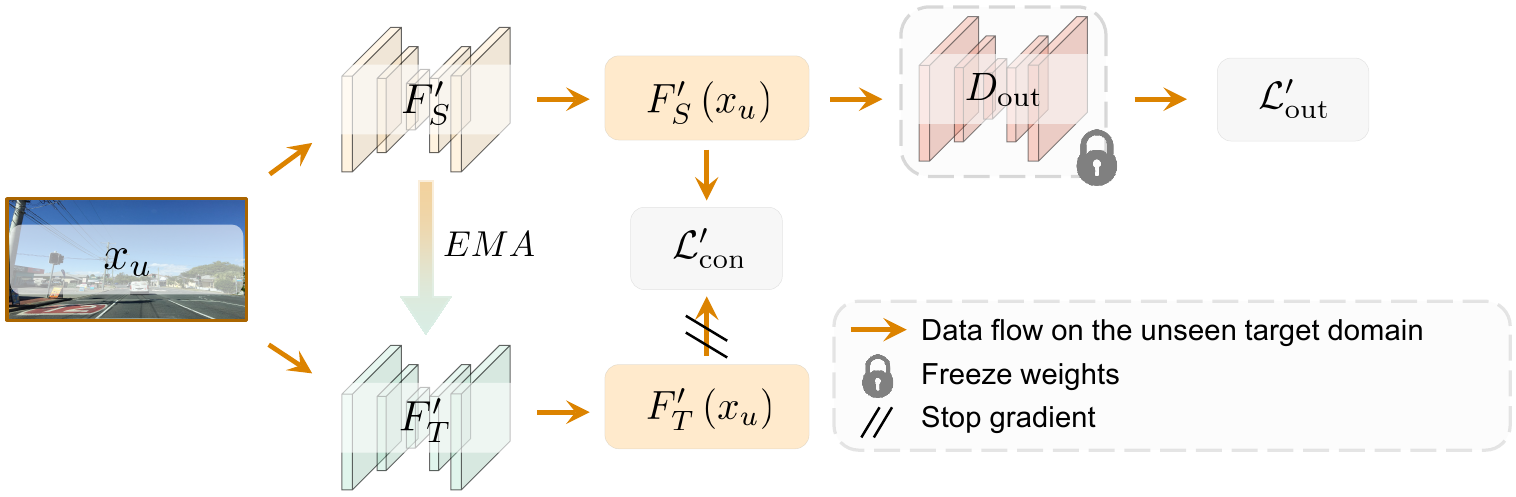}
\caption{Illustration of our \emph{\mta} (\MTA). Given a set of unlabeled mages $X_u$ from an unseen target domain, \MTA distills and adapts the knowledge from a pre-trained \MTL model by self-distillation and one-way adversarial learning. Both student and teacher networks $F'_S, F'_T$ are initialized from the pre-trained model. Same for the discriminator $D_\out$, which remains frozen.}
\label{fig:UTKD}
\end{figure}
%------------------------------------------------------------------------------

\subsection{\Mta}
\label{sec:mta}

Most multi-target UDA approaches for cross-domain semantic segmentation use a predetermined set of target domains \cite{isobe2021multi,saporta2021multi,lee2022adas}. Thus, the learned model still needs to be retrained from scratch on all data when a new unseen target domain is given, which makes it difficult to apply these approaches to new datasets.

To address this challenge, we introduce a \emph{\mta} (\MTA) method that quickly adapts a pre-trained \MTL model to a new unseen target domain without accessing any external data from the source or other target domains. As shown in \autoref{fig:UTKD}, this method is a simplified version of \MTL, where the source-domain data and the supervised loss are removed. The key idea is to perform self-distillation and adversarial learning directly on the new unseen target domain so that the knowledge from the pre-trained \MTL model is distilled and adapted.

To achieve this goal, there are again a student network $F'_S$ and a teacher network $F'_T$. We initialize $F'_T$ according to the pre-trained \MTL model while training $F'_S$ from scratch. At each iteration, $F'_T$ is again obtained by EMA on the parameters of $F'_S$. As in \autoref{sec:mtl}, we perform self-distillation on the unseen target domain data using a consistency loss that minimizes the MSE between the student and teacher predictions
\begin{linenomath}
\begin{align}
	\cL'_\con(X_u, F'_S) &=
		\expect_{x \sim X_u} \ell'_\con(\cA(x), F'_S)
\label{eq:con2-1}
\\
\ell'_\con(x, F'_S) &=
	\frac{1}{K} \sum_{k=1}^K \norm{F'_S(x)^{(k)} - F'_T(x)^{(k)}}^2, \label{eq:con2-2}
\end{align}
\end{linenomath}
where $\cA(x)$ is data augmentation, as in~\eq{con-1}. Although there are no labels in $X_u$, this loss allows the student $F'_S$ to self-train, guided by the teacher $F'_T$.

More importantly, we now also have a pre-trained discriminator $D_\out$ from \MTL that can discriminate segmentation maps between the source and multiple target domains. Considering that the new unseen target domain may be distinctly different from the source domain, the pre-trained $D_\out$ should tend to classify predictions for input examples $x \in X_u$ as the target domain. Since our goal is to make the \MTA model yield domain-invariant segmentation maps on the unseen target domain, a natural idea is to perform adversarial learning to fool the pre-trained $D_\out$ by maximizing the probability of the unknown target-domain predictions being classified as the source-domain. Accordingly, the adversarial loss is
\begin{linenomath}
\begin{equation}
	\cL'_\out(X_u, F'_S) = \cL^-(X_u, F'_S, D_\out),
\label{eq:out2}
\end{equation}
\end{linenomath}
where the negative loss $\cL^-$ is defined in~\eq{adv-neg}. Since there is no source data, there is no positive term as in~\eq{out}. Thus, this is \emph{one-way} adversarial learning. According to our knowledge, we are the first to introduce such an approach in domain adaptation. In addition, we keep the discriminator $D_\out$ fixed, as pre-trained by \MTL. This is what prevents the segmentation model $F'_S$ from forgetting the knowledge acquired from the external data while it is being adapted. There is thus no maximization as in~\eq{mtkd}, and the complete objective function becomes
\begin{linenomath}
\begin{equation}
	\min_{F'_S} \lambda_\con \cL'_\con + \lambda_\out \cL'_\out.
\label{eq:UTKD}
\end{equation}
\end{linenomath}

%------------------------------------------------------------------------------
\begin{figure}
\centering
\fig[1]{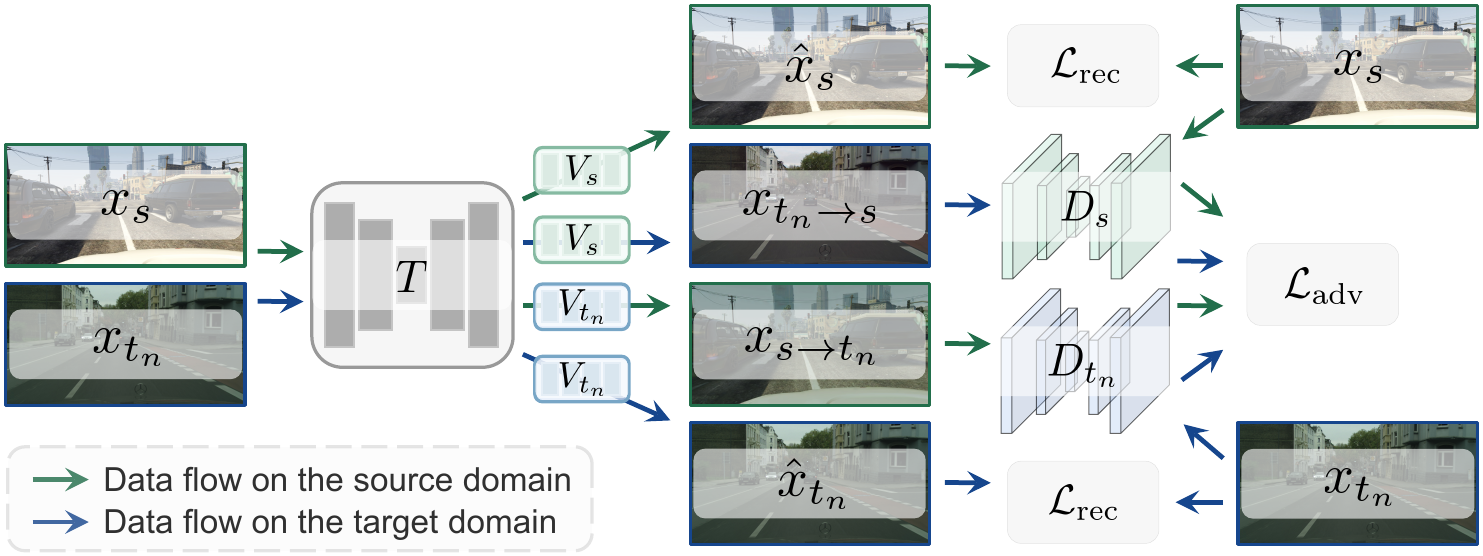}
\caption{Illustration of our \emph{\mts} (\MTS). Given a set of labeled images $X_s$ from the source domain and unlabeled images $\cX_t=\{X_{t_n}\}_{n=1}^N$ from multiple target domains, the style transfer network $T$ learns to either reconstruct, guided by the reconstruction loss $\cL_\rec$, or transfer the style of the input image to another domain, guided by the adversarial loss $\cL_\adv$, depending on the style parameters $V$ that are plugged into $T$ as shown in \autoref{fig:t}. There is one discriminator $D_s, \cD_t=\{D_{t_n}\}_{n=1}^N$ and one set of learnable style parameters $V_s, \cV_t=\{V_{t_n}\}_{n=1}^N$ for each domain. We use $x_{a \to b}$ to denote the transferred image from domain $a$ to $b$. Learning is unsupervised. Only one target domain is shown for brevity.}
\label{fig:mtstn}
\end{figure}
%------------------------------------------------------------------------------

\subsection{\Mts}
\label{sec:mts}

To further mitigate the visual appearance shift between the source and multiple target domains and boost the performance of \MTL, we introduce a \emph{\mts} (\MTS). As shown in \autoref{fig:mtstn}, the main idea is to simultaneously learn the style of each domain. A shared network can then transfer the style from one domain to another, simply by plugging in the target style, while maintaining the content of the original image.

\begin{figure}
\centering
\fig[1]{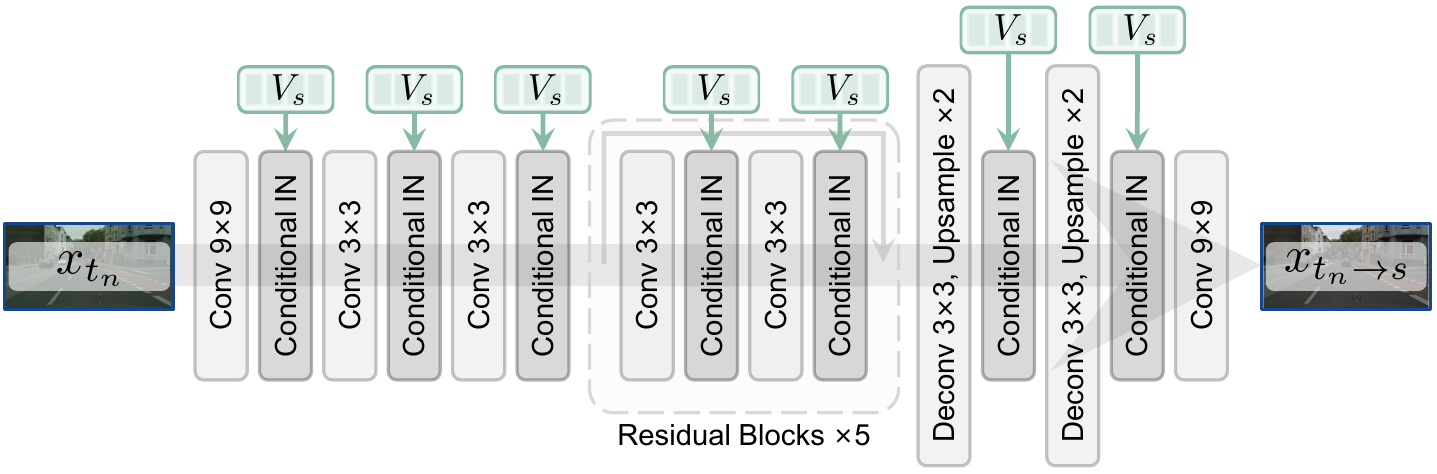}
\caption{Architecture of style transfer network $T$ in our \MTS. Domain style parameters $V$ are plugged into $T$ as parameters of a series of \emph{conditional instance normalization} (CIN) layers. Here, input image $x_{t_n}$ from target domain $X_{t_n}$ is transferred to the style $V_s$ of source domain $X_s$, denoted as $x_{t_n \to s} = T(x_{t_n}, V_s)$. More examples shown in \autoref{fig:mtstn}.}
\label{fig:t}
\end{figure}

Formally, we represent a style as $V = \{\gamma, \beta\}$, where $\gamma, \beta \in \real^d$ are scaling and shifting parameters in a feature space of dimension $d$. We denote by $V_s$ the source domain style and by $\cV_t = \{V_{t_n}\}_{n=1}^N$ the target domain styles. Given a style $V$, the style transfer network $T$ maps an image $x \in [0,1]^K$ to another image $T(x, V) \in [0,1]^K$. We write $T_V(x) = T(x, V)$ for brevity. \autoref{fig:t} illustrates the architecture of $T$, containing a series of \emph{conditional instance normalization} (CIN) layers~\cite{dumoulin2016learned}, all taking the same style as input. Given an intermediate feature map $f$ of $T$, the CIN operation with style $V = \{\gamma, \beta\}$ is defined as
\begin{linenomath}
\begin{equation}
	\cin(f, V) = \gamma \left( \frac{f-\mu(f)}{\sigma(f)} \right) + \beta,
\label{eq:in}
\end{equation}
\end{linenomath}
where $\mu(f)$ and $\sigma(f)$ are the mean and standard deviation over spatial dimensions independently for each channel in $f$, and all operations are element-wise. Importantly, the parameters $V = \{\gamma, \beta\}$ used in~\eq{in} are independent of the network $T$, which can transfer from one style to another simply by switching $V$. The way we learn $\{\gamma, \beta\}$ differs from CIN, which learns each style from a single image, using a style loss on that image~\cite{dumoulin2016learned}. Instead, we aim to learn each style from all training images of one domain, and we achieve this by an adversarial loss.

%------------------------------------------------------------------------------
%------------------------------------------------------------------------------

To maintain the content of the input image for each domain, we define the reconstruction loss
\begin{multline}
	\cL_\rec(X_s,\cX_t,T,V_s,\cV_t) =  \\
 \expect_{x \sim X_s} \ell_\rec(x, T_{V_s}) +
		\sum_{n=1}^N \expect_{x \sim X_{t_n}} \ell_\rec(x, T_{V_{t_n}}),
\label{eq:rec-1}
\end{multline}
where, given an image $x$ and a mapping $F$,
\begin{linenomath}
\begin{align}
	\ell_\rec(x, F) &= \norm{x - F(x)}_1.
\label{eq:rec-2}
\end{align}
\end{linenomath}

To achieve style transfer between different domains, we define a discriminator for each domain. We denote by $D_s$ the discriminator for the source domain and by $\cD_t = \{D_{t_n}\}_{n=1}^N$ the discriminators for the target domains. We then formulate an adversarial loss across domains
\begin{align}
	\cL &_\adv(X_s,\cX_t,T,V_s,\cV_t,D_s,\cD_{t}) =\nonumber \\
	 &\cL^+(X_s,\id,D_s) + \sum_{n=1}^N \cL^-(X_{t_n},T_{V_s},D_s) +\nonumber \\ & \sum_{n=1}^N \left( \cL^+(X_{t_n},\id,D_{t_n}) + \cL^-(X_s,T_{V_{t_n}},D_{t_n}) \right),
\label{eq:adv}
\end{align}
where the positive and negative loss $\cL^+, \cL^-$ are defined in~\eq{adv-pos},~\eq{adv-neg} and $\id$ is the identity function. That is, original images of a domain are treated as positive by the discriminator of that domain (first and third term), while images with style transferred to a domain are treated as negative by the discriminator of that domain (second and fourth term). The complete objective function is
\begin{linenomath}
\begin{equation}
	\min_{T,V_s,\cV_t} \max_{D_s,\cD_t} \cL_\rec + \lambda_\adv \cL_\adv,
\label{eq:mtstn}
\end{equation}
\end{linenomath}
where $\lambda_\adv$ is the weighting factor for the adversarial loss.

\section{Experiments}
\label{sec:exp}

\subsection{Datasets and metrics}
\label{sec:data}

Four benchmark urban driving datasets are adopted in our experiments, including one synthetic dataset (GTA5~\cite{gta}) and three real-world datasets (CityScapes~\cite{cityscapes}, Indian Driving (IDD)~\cite{idd}, and Mapillary~\cite{mapillary}).

\textbf{GTA5} contains 24,966 high-quality labeled frames from the realistic open-world computer games Grand Theft Auto V (GTA5). Each frame is generated from the fictional city Los Santos, based on Los Angeles in Southern California.

\textbf{CityScapes} contains real-world vehicle-egocentric images collected from 50 cities in Germany and its surrounding countries. It is split into training and validation sets of 2,975 and 500 examples respectively.

\textbf{IDD} is a diverse street-view dataset that captures unstructured traffic on roads in India.  It is split into training and validation sets of 6,993 and 981 examples respectively.

\textbf{Mapillary} is a street-view dataset containing high-resolution images collected from all over the world and diverse imaging devices. It is split into training and validation sets of 18,000 and 2,000 examples respectively.

For fair comparisons, we follow the same label mapping protocol used in \cite{saporta2021multi,lee2022adas} and standardize the label set with 7 shared super classes among all four datasets, including \textit{flat}, \textit{construction}, \textit{object}, \textit{nature}, \textit{sky}, \textit{human}, and \textit{vehicle}. When CityScapes, IDD, or Mapillary are used as target domains, only unlabeled images are used at training according to the UDA setting, while the evaluation is conducted with the corresponding labeled validation set.

We report quantitative segmentation results using per-class IoU, mean intersection-over-union (mIoU) over the 7 shared super classes, and the average mIoU over different target domains.

\subsection{Implementation details}
\label{sec:detail}

Following \cite{vu2019AdvEnt,saporta2021multi,lee2022adas}, we use DeepLab-v2~\cite{deeplab} with the ResNet-101~\cite{resnet} network pre-trained on ImageNet~\cite{imagenet} as the segmentation model for both the student $F_S$ and the teacher $F_T$ for fair comparisons. The discriminator $D_\out$ has a DCGAN~\cite{radford2015unsupervised} architecture with $5$ convolutional layers of kernel $4 \times 4$ and stride of $2$. The EMA parameter $\alpha$ in~\eq{ema} is set to $0.999$. The loss factors $\lambda_\con$ and $\lambda_\out$ in \eq{mtkd} and \eq{UTKD} are set to $100$ and $10^{-3}$, respectively.

The data augmentation function $\mathcal{A}$ in~\eq{con-1} is implemented with the CutMix~\cite{french2019semi} strategy and the proposed visual style transfer network $T$. For each target-domain input image $x_{t_n} \in X_{t_n}$, we first transfer its visual style to the source domain with $x_{t_n \to s} = T(x_{t_n}, V_s)$. Then, we use CutMix to generate a mixed example from two transferred target-domain examples. Finally, the teacher predictions for the original two target-domain examples are mixed to produce a pseudo label for the student prediction of the mixed example. For function $\mathcal{A}$ in~\eq{con2-1}, the implementation is the same except that we do not perform visual style transfer on the unseen target domain.

%------------------------------------------------------------------------------
\begin{table}
\centering
\caption{Quantitative cross-domain semantic segmentation results from GTA5 (G) to CityScapes (C) and IDD (I) datasets.}
\resizebox{\linewidth}{!}{
\begin{tabular}{c|ccc|ccccccc|cc} %
\toprule
\Th{Method}&\Th{Flow}&\rotatebox{90}{\Th{Target}}&\rotatebox{90}{\Th{Extern}}&\rotatebox{90}{\emph{flat}}&\rotatebox{90}{\emph{constr.}}&\rotatebox{90}{\emph{object}}&\rotatebox{90}{\emph{nature}}&\rotatebox{90}{\emph{sky}}&\rotatebox{90}{\emph{human}}&\rotatebox{90}{\emph{vehicle}}&\Th{mIoU}&\Th{Avg}\\
\midrule
URMA  \cite{Teja2021uncertainty}&G$\to$C&C&\xmark&91.1&78.9&26.1&80.7&74.6&60.9&67.7&68.6&\multirow{2}{*}{67.1}\\
(source-free)&G$\to$I&I&\xmark&93.0&52.9&15.8&78.5&90.4&54.8&74.6&65.7&\\
\midrule
AdvStyle  \cite{zhong2022adversarial}&G$\to$C&C&\xmark&87.2&71.8&25.5&82.2&81.0&59.9&79.2&69.5&\multirow{2}{*}{67.2}\\
(domain gen.)&G$\to$I&I&\xmark&88.2&49.9&13.4&77.9&90.9&55.9&78.5&64.9&\\
\midrule
&G$\to$C&C&\cmark&93.5&80.5&26.0&78.5&78.5&55.1&76.4&69.8 (*)&\multirow{2}{*}{66.5}\\
AdvEnt \cite{vu2019AdvEnt}&G$\to$C&I&\xmark&91.3&52.3&13.3&76.1&88.7&46.7&74.8&63.3\textcolor[rgb]{1,0,0}{$_{\downarrow 1.8}$}&\\
(single-target)&G$\to$I&C&\xmark&78.6&79.2&24.8&77.6&83.6&48.7&44.8&62.5\textcolor[rgb]{1,0,0}{$_{\downarrow 7.3}$}&\multirow{2}{*}{63.8}\\
&G$\to$I&I&\cmark&91.2&53.1&16.0&78.2&90.7&47.9&78.9&65.1 (*)&\\
\midrule
\rowcolor{LightCyan}
&G$\to$C&C&\cmark&95.9&\textbf{85.5}&40.2&84.8&81.4&64.1&82.2&76.3\textcolor[rgb]{0,0.5,0}{$_{\uparrow 6.5}$}&\\
\rowcolor{LightCyan}
\MTL&G$\to$C&I&\xmark&92.5&58.3&19.2&79.3&91.8&56.9&81.6&68.5\textcolor[rgb]{0,0.5,0}{$_{\uparrow 3.4}$}&\multirow{-2}{*}{72.4}\\
\rowcolor{LightCyan}
(single-target)&G$\to$I&C&\xmark&95.3&83.7&35.9&83.9&78.5&64.7&79.9&74.5\textcolor[rgb]{0,0.5,0}{$_{\uparrow 4.7}$}&\\
\rowcolor{LightCyan}
&G$\to$I&I&\cmark&\textbf{94.2}&58.3&\textbf{25.0}&\textbf{82.9}&\textbf{92.8}&\textbf{61.6}&\textbf{85.3}&\textbf{71.4}\textcolor[rgb]{0,0.5,0}{$_{\uparrow 6.3}$}&\multirow{-2}{*}{72.9}\\
\midrule
AdvEnt \cite{vu2019AdvEnt}&G$\to$\{C, I\}&C&\cmark&93.9&80.2&26.2&79.0&80.5&52.5&78.0&70.0\textcolor[rgb]{0,0.5,0}{$_{\uparrow 0.2}$}&\multirow{2}{*}{67.4}\\
(multi-target)&G$\to$\{C, I\}&I&\cmark&91.8&54.5&14.4&76.8&90.3&47.5&78.3&64.8\textcolor[rgb]{1,0,0}{$_{\downarrow 0.3}$}&\\
\midrule
MTKT \cite{saporta2021multi}&G$\to$\{C, I\}&C&\cmark&94.5&82.0&23.7&80.1&\textbf{84.0}&51.0&77.6&70.4\textcolor[rgb]{0,0.5,0}{$_{\uparrow 0.6}$}&\multirow{2}{*}{68.2}\\
(multi-target)&G$\to$\{C, I\}&I&\cmark&91.4&56.6&13.2&77.3&91.4&51.4&79.9&65.9\textcolor[rgb]{0,0.5,0}{$_{\uparrow 0.8}$}&\\
\midrule
ADAS \cite{lee2022adas}&G$\to$\{C, I\}&C&\cmark&95.1&82.6&39.8&84.6&81.2&63.6&80.7&75.4\textcolor[rgb]{0,0.5,0}{$_{\uparrow 5.6}$}&\multirow{2}{*}{71.2}\\
(multi-target)&G$\to$\{C, I\}&I&\cmark&90.5&\textbf{63.0}&22.2&73.7&87.9&54.3&76.9&66.9\textcolor[rgb]{0,0.5,0}{$_{\uparrow 1.8}$}&\\
\midrule
\rowcolor{LightCyan}
\MTL&G$\to$\{C, I\}&C&\cmark&96.2&85.3&40.3&\textbf{85.1}&80.1&\textbf{65.2}&83.6&76.5\textcolor[rgb]{0,0.5,0}{$_{\uparrow 6.7}$}&\\
\rowcolor{LightCyan}
(multi-target)&G$\to$\{C, I\}&I&\cmark&94.1&60.3&23.2&82.7&92.7&60.3&\textbf{85.3}&71.2\textcolor[rgb]{0,0.5,0}{$_{\uparrow 6.1}$}&\multirow{-2}{*}{\textbf{73.8}}\\
\midrule
\rowcolor{LightCyan}
\MTA&(G$\to$I)$\to$C&C&\xmark&\textbf{97.0}&84.7&\textbf{41.2}&\textbf{85.1}&81.8&64.3&\textbf{85.2}&\textbf{77.0}\textcolor[rgb]{0,0.5,0}{$_{\uparrow 7.2}$}&\\
\rowcolor{LightCyan}
(multi-target)&(G$\to$C)$\to$I&I&\xmark&92.7&59.1&24.5&79.3&91.9&61.0&85.0&70.5\textcolor[rgb]{0,0.5,0}{$_{\uparrow 5.4}$}&\multirow{-2}{*}{73.7}\\
\bottomrule
\end{tabular}
}
\\\vspace{1pt}
\raggedright \scriptsize \justifying \noindent \textbf{Bold}: best IoU (\%) over all methods in each target domain. \textcolor[rgb]{0,0.5,0}{Green} / \textcolor[rgb]{1,0,0}{red}: mIoU gain / loss \wrt the corresponding per-target baseline, marked by `*'. \Th{Extern}: using external data from the source or other target domains.
\label{tab:ci}
\end{table}
%------------------------------------------------------------------------------

%------------------------------------------------------------------------------
\begin{table}
\centering
\caption{Quantitative cross-domain semantic segmentation results from GTA5 (G) to CityScapes (C) and Mapillary (M) datasets.}
\resizebox{\linewidth}{!}{
\begin{tabular}{c|ccc|ccccccc|cc} %
\toprule
\Th{Method}&\Th{Flow}&\rotatebox{90}{\Th{Target}}&\rotatebox{90}{\Th{Extern}}&\rotatebox{90}{\emph{flat}}&\rotatebox{90}{\emph{constr.}}&\rotatebox{90}{\emph{object}}&\rotatebox{90}{\emph{nature}}&\rotatebox{90}{\emph{sky}}&\rotatebox{90}{\emph{human}}&\rotatebox{90}{\emph{vehicle}}&\Th{mIoU}&\Th{Avg}\\
\midrule
URMA  \cite{Teja2021uncertainty}&G$\to$C&C&\xmark&91.1&78.9&26.1&80.7&74.6&60.9&67.7&68.6&\multirow{2}{*}{69.5}\\
(source-free)&G$\to$M&M&\xmark&88.3&71.3&39.0&72.9&90.4&56.5&74.5&70.4&\\
\midrule
AdvStyle  \cite{zhong2022adversarial}&G$\to$C&C&\xmark&87.2&71.8&25.5&82.2&81.0&59.9&79.2&69.5&\multirow{2}{*}{70.2}\\
(domain gen.)&G$\to$M&M&\xmark&87.5&70.9&33.4&72.8&90.9&62.1&79.1&70.9&\\
\midrule
&G$\to$C&C&\cmark&93.5&80.5&26.0&78.5&78.5&55.1&76.4&69.8 (*)&\multirow{2}{*}{66.6}\\
AdvEnt \cite{vu2019AdvEnt}&G$\to$C&M&\xmark&86.8&69.0&30.2&71.2&91.5&35.3&59.5&63.4\textcolor[rgb]{1,0,0}{$_{\downarrow 6.2}$}&\\
(single-target)&G$\to$M&C&\xmark&89.3&79.3&19.5&76.9&84.6&47.7&63.0&65.8\textcolor[rgb]{1,0,0}{$_{\downarrow 4.0}$}&\multirow{2}{*}{67.7}\\
&G$\to$M&M&\cmark&89.5&72.6&31.0&75.3&94.1&50.7&73.8&69.6 (*)&\\
\midrule
\rowcolor{LightCyan}
&G$\to$C&C&\cmark&95.9&85.5&40.2&84.8&81.4&64.1&82.2&76.3\textcolor[rgb]{0,0.5,0}{$_{\uparrow 6.5}$}&\\
\rowcolor{LightCyan}
\MTL&G$\to$C&M&\xmark&89.7&76.2&44.1&75.5&94.1&63.0&83.3&75.1\textcolor[rgb]{0,0.5,0}{$_{\uparrow 5.5}$}&\multirow{-2}{*}{75.7}\\
\rowcolor{LightCyan}
(single-target)&G$\to$M&C&\xmark&\textbf{96.6}&84.5&37.7&84.7&80.5&61.8&82.8&75.5\textcolor[rgb]{0,0.5,0}{$_{\uparrow 5.7}$}&\\
\rowcolor{LightCyan}
&G$\to$M&M&\cmark&90.0&76.4&\textbf{47.5}&74.1&93.7&60.1&\textbf{84.6}&75.2\textcolor[rgb]{0,0.5,0}{$_{\uparrow 5.6}$}&\multirow{-2}{*}{75.3}\\
\midrule
AdvEnt \cite{vu2019AdvEnt}&G$\to$\{C, M\}&C&\cmark&93.1&80.5&24.0&77.9&81.0&52.5&75.0&69.1\textcolor[rgb]{1,0,0}{$_{\downarrow 0.7}$}&\multirow{2}{*}{68.9}\\
(multi-target)&G$\to$\{C, M\}&M&\cmark&90.0&71.3&31.1&73.0&92.6&46.6&76.6&68.7\textcolor[rgb]{1,0,0}{$_{\downarrow 0.9}$}&\\
\midrule
MTKT \cite{saporta2021multi}&G$\to$\{C, M\}&C&\cmark&95.0&81.6&23.6&80.1&83.6&53.7&79.8&71.1\textcolor[rgb]{0,0.5,0}{$_{\uparrow 1.3}$}&\multirow{2}{*}{70.9}\\
(multi-target)&G$\to$\{C, M\}&M&\cmark&\textbf{90.6}&73.3&31.0&75.3&\textbf{94.5}&52.2&79.8&70.8\textcolor[rgb]{0,0.5,0}{$_{\uparrow 1.2}$}&\\
\midrule
ADAS \cite{lee2022adas}&G$\to$\{C, M\}&C&\cmark&96.4&83.5&35.1&83.8&\textbf{84.9}&62.3&81.3&75.3\textcolor[rgb]{0,0.5,0}{$_{\uparrow 5.5}$}&\multirow{2}{*}{73.9}\\
(multi-target)&G$\to$\{C, M\}&M&\cmark&88.6&73.7&41.0&75.4&93.4&58.5&77.2&72.6\textcolor[rgb]{0,0.5,0}{$_{\uparrow 3.0}$}&\\
\midrule
\rowcolor{LightCyan}
\MTL&G$\to$\{C, M\}&C&\cmark&96.3&\textbf{85.6}&39.8&\textbf{85.5}&82.5&\textbf{64.5}&\textbf{83.5}&\textbf{76.8}\textcolor[rgb]{0,0.5,0}{$_{\uparrow 7.0}$}&\\
\rowcolor{LightCyan}
(multi-target)&G$\to$\{C, M\}&M&\cmark&89.9&\textbf{76.7}&46.3&73.5&93.2&\textbf{63.8}&84.1&\textbf{75.3}\textcolor[rgb]{0,0.5,0}{$_{\uparrow 5.7}$}&\multirow{-2}{*}{\textbf{76.0}}\\
\midrule
\rowcolor{LightCyan}
\MTA&(G$\to$M)$\to$C&C&\xmark&\textbf{96.6}&84.7&\textbf{43.1}&85.4&82.8&62.6&82.9&\textbf{76.8}\textcolor[rgb]{0,0.5,0}{$_{\uparrow 7.0}$}&\\
\rowcolor{LightCyan}
(multi-target)&(G$\to$C)$\to$M&M&\xmark&90.1&75.2&46.7&\textbf{76.2}&94.4&60.1&82.9&75.1\textcolor[rgb]{0,0.5,0}{$_{\uparrow 5.5}$}&\multirow{-2}{*}{75.9}\\
\bottomrule
\end{tabular}
}
\\\vspace{1pt}
\raggedright \scriptsize \justifying \noindent \textbf{Bold}: best IoU (\%) over all methods in each target domain.  \textcolor[rgb]{0,0.5,0}{Green} / \textcolor[rgb]{1,0,0}{red}: mIoU gain / loss \wrt the corresponding per-target baseline, marked by `*'. \Th{Extern}: using external data from the source or other target domains.
\label{tab:cm}
\end{table}
%------------------------------------------------------------------------------

%------------------------------------------------------------------------------
\begin{table}
\centering
\caption{Quantitative cross-domain semantic segmentation results from GTA5 (G) to CityScapes (C), IDD (I), and Mapillary (M) datasets.}
\resizebox{\linewidth}{!}{
\begin{tabular}{c|ccc|ccccccc|cc} %
\toprule
\Th{Method}&\Th{Flow}&\rotatebox{90}{\Th{Target}}&\rotatebox{90}{\Th{Extern}}&\rotatebox{90}{\emph{flat}}&\rotatebox{90}{\emph{constr.}}&\rotatebox{90}{\emph{object}}&\rotatebox{90}{\emph{nature}}&\rotatebox{90}{\emph{sky}}&\rotatebox{90}{\emph{human}}&\rotatebox{90}{\emph{vehicle}}&\Th{mIoU}&\Th{Avg}\\
\midrule
\multirow{3}{*}{\shortstack{URMA \cite{Teja2021uncertainty}\\(source-free)}}&G$\to$C&C&\xmark&91.1&78.9&26.1&80.7&74.6&60.9&67.7&68.6&\multirow{3}{*}{68.2}\\
&G$\to$I&I&\cmark&93.0&52.9&15.8&78.5&90.4&54.8&74.6&65.7&\\
&G$\to$M&M&\xmark&88.3&71.3&39.0&72.9&90.4&56.5&74.5&70.4&\\
\midrule
\multirow{3}{*}{\shortstack{AdvStyle \cite{zhong2022adversarial}\\(domain gen.)}}&G$\to$C&C&\xmark&87.2&71.8&25.5&82.2&81.0&59.9&79.2&69.5&\multirow{3}{*}{68.4}\\
&G$\to$I&I&\cmark&88.2&49.9&13.4&77.9&90.9&55.9&78.5&64.9&\\
&G$\to$M&M&\xmark&87.5&70.9&33.4&72.8&90.9&62.1&79.1&70.9&\\
\midrule
&G$\to$C&C&\cmark&93.5&80.5&26.0&78.5&78.5&55.1&76.4&69.8 (*)&\multirow{3}{*}{65.5}\\
&G$\to$C&I&\xmark&91.3&52.3&13.3&76.1&88.7&46.7&74.8&63.3\textcolor[rgb]{1,0,0}{$_{\downarrow 1.8}$}&\\
&G$\to$C&M&\xmark&86.8&69.0&30.2&71.2&91.5&35.3&59.5&63.4\textcolor[rgb]{1,0,0}{$_{\downarrow 6.2}$}&\\
\multirow{3}{*}{\shortstack{AdvEnt \cite{vu2019AdvEnt}\\(single-target)}}&G$\to$I&C&\xmark&78.6&79.2&24.8&77.6&83.6&48.7&44.8&62.5\textcolor[rgb]{1,0,0}{$_{\downarrow 7.3}$}&\multirow{3}{*}{65.5}\\
&G$\to$I&I&\cmark&91.2&53.1&16.0&78.2&90.7&47.9&78.9&65.1 (*)&\\
&G$\to$I&M&\xmark&88.5&71.2&32.4&72.8&92.8&51.3&73.7&69.0\textcolor[rgb]{1,0,0}{$_{\downarrow 0.6}$}&\\
&G$\to$M&C&\xmark&89.3&79.3&19.5&76.9&84.6&47.7&63.0&65.8\textcolor[rgb]{1,0,0}{$_{\downarrow 4.0}$}&\multirow{3}{*}{66.7}\\
&G$\to$M&I&\xmark&91.7&54.3&13.0&77.3&92.3&47.4&76.8&64.7\textcolor[rgb]{1,0,0}{$_{\downarrow 0.4}$}&\\
&G$\to$M&M&\cmark&89.5&72.6&31.0&75.3&94.1&50.7&73.8&69.6 (*)&\\
\midrule
\rowcolor{LightCyan}
&G$\to$C&C&\cmark&95.9&85.5&40.2&84.8&81.4&64.1&82.2&76.3\textcolor[rgb]{0,0.5,0}{$_{\uparrow 6.5}$}&\\
\rowcolor{LightCyan}
&G$\to$C&I&\xmark&92.5&58.3&19.2&79.3&91.8&56.9&81.6&68.5\textcolor[rgb]{0,0.5,0}{$_{\uparrow 3.4}$}&\\
\rowcolor{LightCyan}
&G$\to$C&M&\xmark&89.7&76.2&44.1&75.5&94.1&63.0&83.3&75.1\textcolor[rgb]{0,0.5,0}{$_{\uparrow 5.5}$}&\multirow{-3}{*}{73.3}\\
\rowcolor{LightCyan}
&G$\to$I&C&\xmark&95.3&83.7&35.9&83.9&78.5&64.7&79.9&74.5\textcolor[rgb]{0,0.5,0}{$_{\uparrow 4.7}$}&\\
\rowcolor{LightCyan}
&G$\to$I&I&\cmark&94.2&58.3&25.0&82.9&\textbf{92.8}&\textbf{61.6}&\textbf{85.3}&71.4\textcolor[rgb]{0,0.5,0}{$_{\uparrow 6.3}$}&\\
\rowcolor{LightCyan}
\multirow{-3}{*}{\shortstack{\MTL\\(single-target)}}&G$\to$I&M&\xmark&89.9&75.6&42.9&74.7&93.8&60.8&82.6&74.3\textcolor[rgb]{0,0.5,0}{$_{\uparrow 4.7}$}&\multirow{-3}{*}{73.4}\\
\rowcolor{LightCyan}
&G$\to$M&C&\xmark&96.6&84.5&37.7&84.7&80.5&61.8&82.8&75.5\textcolor[rgb]{0,0.5,0}{$_{\uparrow 5.7}$}&\\
\rowcolor{LightCyan}
&G$\to$M&I&\xmark&94.4&58.1&26.1&81.6&92.2&56.8&81.7&70.1\textcolor[rgb]{0,0.5,0}{$_{\uparrow 5.0}$}&\\
\rowcolor{LightCyan}
&G$\to$M&M&\cmark&90.0&76.4&\textbf{47.5}&74.1&93.7&60.1&\textbf{84.6}&75.2\textcolor[rgb]{0,0.5,0}{$_{\uparrow 5.6}$}&\multirow{-3}{*}{73.6}\\
\midrule
\multirow{3}{*}{\shortstack{AdvEnt \cite{vu2019AdvEnt}\\(multi-target)}}&G$\to$\{C, I, M\}&C&\cmark&93.6&80.6&26.4&78.1&81.5&51.9&76.4&69.8 --&\multirow{3}{*}{67.8}\\
&G$\to$\{C, I, M\}&I&\cmark&92.0&54.6&15.7&77.2&90.5&50.8&78.6&65.6\textcolor[rgb]{1,0,0}{$_{\uparrow 0.5}$}&\\
&G$\to$\{C, I, M\}&M&\cmark&89.2&72.4&32.4&73.0&92.7&41.6&74.9&68.0\textcolor[rgb]{1,0,0}{$_{\downarrow 1.6}$}&\\
\midrule
\multirow{3}{*}{\shortstack{MTKT \cite{saporta2021multi}\\(multi-target)}}&G$\to$\{C, I, M\}&C&\cmark&94.6&80.7&23.8&79.0&84.5&51.0&79.2&70.4\textcolor[rgb]{0,0.5,0}{$_{\uparrow 0.6}$}&\multirow{3}{*}{69.1}\\
&G$\to$\{C, I, M\}&I&\cmark&91.7&55.6&14.5&78.0&92.6&49.8&79.4&65.9\textcolor[rgb]{0,0.5,0}{$_{\uparrow 0.8}$}&\\
&G$\to$\{C, I, M\}&M&\cmark&\textbf{90.5}&73.7&32.5&75.5&94.3&51.2&80.2&71.1\textcolor[rgb]{0,0.5,0}{$_{\uparrow 1.5}$}&\\
\midrule
\multirow{3}{*}{\shortstack{ADAS \cite{lee2022adas}\\(multi-target)}}&G$\to$\{C, I, M\}&C&\cmark&95.8&82.4&38.3&82.4&\textbf{85.0}&60.5&80.2&74.9\textcolor[rgb]{0,0.5,0}{$_{\uparrow 5.1}$}&\multirow{3}{*}{71.3}\\
&G$\to$\{C, I, M\}&I&\cmark&89.9&52.7&25.0&78.1&92.1&51.0&77.9&66.7\textcolor[rgb]{0,0.5,0}{$_{\uparrow 1.6}$}&\\
&G$\to$\{C, I, M\}&M&\cmark&89.2&71.5&45.2&75.8&92.3&56.1&75.4&72.2\textcolor[rgb]{0,0.5,0}{$_{\uparrow 2.6}$}&\\
\midrule
\rowcolor{LightCyan}
&G$\to$\{C, I, M\}&C&\cmark&95.3&\textbf{85.6}&39.7&84.5&82.3&\textbf{65.5}&81.4&76.3\textcolor[rgb]{0,0.5,0}{$_{\uparrow 6.5}$}&\\
\rowcolor{LightCyan}
&G$\to$\{C, I, M\}&I&\cmark&93.9&\textbf{59.7}&22.8&82.1&92.7&60.3&84.6&70.8\textcolor[rgb]{0,0.5,0}{$_{\uparrow 5.7}$}&\\
\rowcolor{LightCyan}
\multirow{-3}{*}{\shortstack{\MTL\\(multi-target)}}&G$\to$\{C, I, M\}&M&\cmark&89.9&\textbf{76.5}&46.9&73.4&93.2&63.8&84.2&75.4\textcolor[rgb]{0,0.5,0}{$_{\uparrow 5.8}$}&\multirow{-3}{*}{74.1}\\
\midrule
\rowcolor{LightCyan}
&G$\to$\{I, M\}&C&\xmark&96.7&84.3&38.2&84.7&78.9&64.6&84.3&75.9\textcolor[rgb]{0,0.5,0}{$_{\uparrow 6.1}$}&\\
\rowcolor{LightCyan}
&G$\to$\{C, M\}&I&\xmark&93.9&58.6&22.7&81.4&91.7&57.7&82.0&69.7\textcolor[rgb]{0,0.5,0}{$_{\uparrow 4.6}$}&\\
\rowcolor{LightCyan}
\multirow{-3}{*}{\shortstack{\MTL$\!^\dagger$\\(multi-target)}}&G$\to$\{C, I\}&M&\xmark&89.7&76.3&44.1&75.4&94.1&63.0&83.4&75.1\textcolor[rgb]{0,0.5,0}{$_{\uparrow 5.5}$}&\multirow{-3}{*}{73.5}\\
\midrule
\rowcolor{LightCyan}
&(G$\to$\{I, M\})$\to$C&C&\xmark&\textbf{97.0}&85.0&\textbf{41.7}&\textbf{85.5}&81.9&65.1&\textbf{84.9}&\textbf{77.3}\textcolor[rgb]{0,0.5,0}{$_{\uparrow 7.5}$}&\\
\rowcolor{LightCyan}
&(G$\to$\{C, M\})$\to$I&I&\xmark&\textbf{95.0}&58.9&\textbf{30.6}&\textbf{83.8}&91.5&60.7&85.0&\textbf{72.2}\textcolor[rgb]{0,0.5,0}{$_{\uparrow 7.1}$}&\\
\rowcolor{LightCyan}
\multirow{-3}{*}{\shortstack{\MTA\\(multi-target)}}&(G$\to$\{C, I\})$\to$M&M&\xmark&89.8&74.0&46.4&\textbf{76.6}&\textbf{94.4}&\textbf{64.5}&84.2&\textbf{75.7}\textcolor[rgb]{0,0.5,0}{$_{\uparrow 6.1}$}&\multirow{-3}{*}{\textbf{75.0}}\\
\bottomrule
\end{tabular}
}
\\\vspace{1pt}
\raggedright \scriptsize \justifying \noindent \textbf{Bold}: best IoU (\%) over all methods in each target domain.  \textcolor[rgb]{0,0.5,0}{Green} / \textcolor[rgb]{1,0,0}{red}: mIoU gain / loss \wrt the corresponding per-target baseline, marked by `*'. \Th{Extern}: using external data from the source or other target domains. \MTL$\!^\dagger$: direct transfer from a pre-trained \MTL model to an unseen target domain.
\label{tab:cim}
\end{table}
%------------------------------------------------------------------------------

\subsection{Synthetic-to-real adaptation}
\label{sec:s2r}

We first evaluate the performance of the proposed methods against existing approaches in the synthetic-to-real adaptation scenario, where the labeled GTA5 dataset is adopted as the source domain and the unlabeled CityScapes, IDD, and Mapillary datasets are used as the multi-target domains. Results are reported in Tables~\ref{tab:ci}, \ref{tab:cm}, and \ref{tab:cim}. It can be observed that AdvEnt~\cite{vu2019AdvEnt} trained with the single-target domain adaptation setting generally yields lower mIoU scores compared to its counterpart in the multi-target domain adaptation setting, which demonstrates the advantage of incorporating multi-domain data into training.

\begin{figure}
\centering
\fig{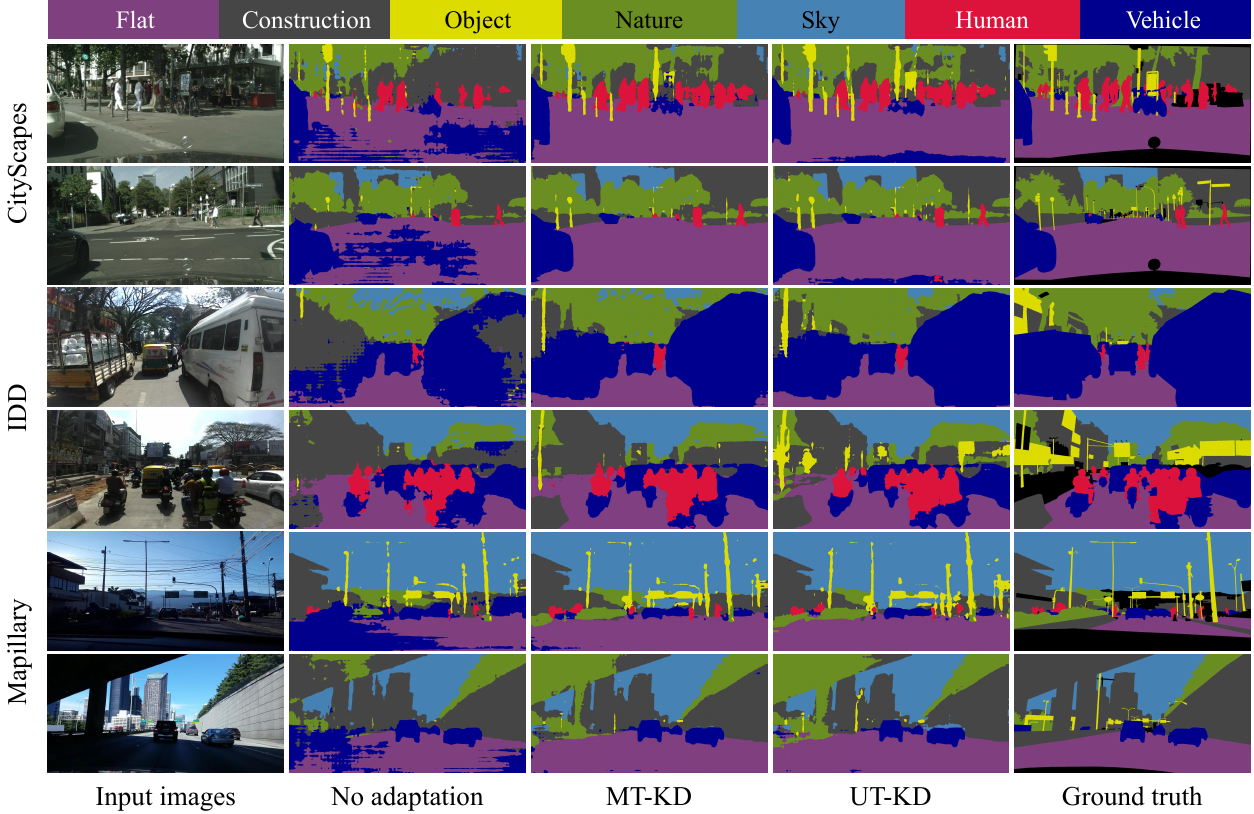}
\caption{Qualitative cross-domain semantic segmentation results from GTA5 to CityScapes, IDD, and Mapillary datasets.}
\label{fig:segmap}
\end{figure}

In all multi-target domain adaptation scenarios, the proposed method \MTL obtains the highest mIoU scores and outperforms the existing state-of-the-art methods like ADAS~\cite{lee2022adas} by a large margin, more than 2\% mIoU. Qualitative adaptation results of \MTL from GTA5 to CityScapes, IDD, and Mapillary are shown in \autoref{fig:segmap}.

Another intriguing finding is that the proposed method \MTA yields very competitive performance compared to \MTL, although it does not access any external data. In \autoref{tab:cim} for example, \MTL yields $74.1\%$ averaged mIoU and \MTA yields $75.0\%$, even outperforming \MTL by $0.9\%$ and ADAS~\cite{lee2022adas} by $3.7\%$. In \autoref{tab:ci} and \autoref{tab:cm}, it is nearly on par with \MTL, losing only by $0.1\%$, and still outperforms ADAS~\cite{lee2022adas} by $2.5\%$ and $2\%$ respectively. Considering that in real-world scenarios, it is more common to get access to pre-trained models than to complete street-view datasets collected from other cities because of data privacy, our \MTA is more flexible and more practical without losing on performance.

%------------------------------------------------------------------------------
%------------------------------------------------------------------------------

\section{Conclusion}

In this paper, we introduce a new strategy for conducting multi-target unsupervised domain adaptation for semantic segmentation without relying on external data. To implement this idea, we first propose the \mtl (\MTL) method, which achieves multi-target UDA for semantic segmentation using adversarial learning and self-distillation, setting new state-of-the-art performance.
As a simplified version, we further propose the \mta (\MTA) method, which rapidly adapts a pre-trained \MTL model to a new unseen target domain through ``one-way'' adversarial learning, without accessing any external data from the source or other target domains. Despite its simplicity, \MTA is more scalable than existing multi-target UDA solutions in handling unseen domains, especially under data privacy constraints. It does not compromise performance compared to \MTL and still outperforms other state-of-the-art methods. To further address the visual appearance shift, we perform visual style transfer across multiple domains by parameterizing the style of each domain through a single vector, thus decoupling it from the style transfer process itself. The latter is accomplished by a \mts (\MTS), which is shared across all domains.

Although the proposed methods are originally designed for the cross-domain semantic segmentation task, they may also be helpful for solving other cross-domain tasks. We will explore it in future work.

\section*{Acknowledgment}

The authors would like to thank the Institute of Advanced Research in Artificial Intelligence (IARAI) for its support.

%%%%%%%%% REFERENCES
{\small
\bibliographystyle{ieee_fullname}
\bibliography{UTKD}
}

\clearpage
\appendix
\noindent{\Large \textbf{Supplementary Material}}

\vspace{.5em}
In this supplementary material, we provide more training details as well as quantitative and qualitative results.

\section{Training details}

We use stochastic gradient descent (SGD) with a learning rate of $2.5 \times 10^{-5}$ to train $F_S$, while $D_\out$ is trained by the Adam optimizer~\cite{kingma2014adam} with a learning rate of $10^{-5}$ and $\beta_1 = 0.9$, $\beta_2 = 0.99$. For both optimizers, we set a weight decay of $5 \times 10^{-5}$ and adopt the ``poly'' learning rate decay schedule, where the initial learning rate is multiplied by $(1-i/I)^p$ with $p = 0.9$, where $i$ is the current iteration and $I$ the total number of iterations, set to $50,000$.

To train \MTS, we use the Adam optimizer for 20 training epochs with weight decay $5 \times 10^{-5}$ and learning rate $2.5 \times 10^{-4}$ and $10^{-5}$ for the generator and discriminators, respectively. Each mini-batch consists of one source-domain image and one target-domain image. The loss factor $\lambda_\adv$ in~\eq{adv} is empirically set to $10^{-3}$.

At inference, we use the teacher network $F_T$ for \MTL and $F'_T$ for \MTA as obtained at the end of training to perform semantic segmentation of input test images.

%------------------------------------------------------------------------------

\section{Real-to-real adaptation}
\label{sec:r2r}

%------------------------------------------------------------------------------

We further conduct experiments on a real-to-real adaptation setting, where the labeled dataset CityScapes is adopted as the source domain and the unlabeled datasets IDD and Mapillary are used as the multi-target domains. As can be observed from \autoref{tab:im}, our \MTL achieves again the best multi-target domain adaptation performance compared to existing approaches. \MTA yields $74.6\%$ averaged mIoU, which is again higher than \MTL by $0.7\%$ and ADAS~\cite{lee2022adas} by $1.9\%$, despite not having access to external data. These results confirm the high practical value of our new UDA strategy without external data.

%------------------------------------------------------------------------------
\begin{table}
\centering
\caption{Quantitative cross-domain semantic segmentation results from CityScapes (C) to IDD (I) and Mapillary (M) datasets.}
\resizebox{\linewidth}{!}{
\begin{tabular}{c|ccc|ccccccc|cc} %
\toprule
\Th{Method}&\Th{Flow}&\rotatebox{90}{\Th{Target}}&\rotatebox{90}{\Th{Extern}}&\rotatebox{90}{\emph{flat}}&\rotatebox{90}{\emph{constr.}}&\rotatebox{90}{\emph{object}}&\rotatebox{90}{\emph{nature}}&\rotatebox{90}{\emph{sky}}&\rotatebox{90}{\emph{human}}&\rotatebox{90}{\emph{vehicle}}&\Th{mIoU}&\Th{Avg}\\
\midrule
URMA  \cite{Teja2021uncertainty}&C$\to$I&I&\xmark&93.9&56.0&23.4&83.7&93.6&52.0&79.2&68.8&\multirow{2}{*}{68.4}\\
(source-free)&C$\to$M&M&\xmark&88.1&71.6&26.5&70.8&92.2&56.5&70.9&68.1&\\
\midrule
AdvStyle  \cite{zhong2022adversarial}&C$\to$I&I&\xmark&93.9&52.9&18.6&82.9&92.6&51.2&76.9&67.0&\multirow{2}{*}{68.8}\\
(domain gen.)&C$\to$M&M&\xmark&89.5&70.2&34.4&77.3&93.1&56.6&73.7&70.6&\\
\midrule
&C$\to$I&I&\cmark&93.2&53.4&16.5&83.4&93.4&51.4&79.5&67.3 (*)&\multirow{2}{*}{68.0}\\
AdvEnt \cite{vu2019AdvEnt}&C$\to$I&M&\xmark&88.2&70.0&28.5&75.4&93.6&49.1&76.7&68.8\textcolor[rgb]{0,0.5,0}{$_{\uparrow 2.2}$}&\\
(single-target)&C$\to$M&I&\xmark&91.8&52.2&15.9&80.2&91.1&45.7&77.6&65.0\textcolor[rgb]{1,0,0}{$_{\downarrow 2.3}$}&\multirow{2}{*}{65.8}\\
&C$\to$M&M&\cmark&87.4&65.9&28.2&72.8&92.1&46.9&72.7&66.6 (*)&\\
\midrule
\rowcolor{LightCyan}
&C$\to$I&I&\cmark&93.7&59.2&29.8&83.6&93.3&62.1&\textbf{85.3}&72.4\textcolor[rgb]{0,0.5,0}{$_{\uparrow 5.1}$}&\\
\rowcolor{LightCyan}
\MTL&C$\to$I&M&\xmark&90.3&75.0&46.2&77.6&94.2&\textbf{63.9}&82.3&75.6\textcolor[rgb]{0,0.5,0}{$_{\uparrow 9.0}$}&\multirow{-2}{*}{74.0}\\
\rowcolor{LightCyan}
(single-target)&C$\to$M&I&\xmark&95.1&58.0&28.7&84.8&92.6&57.7&81.8&71.2\textcolor[rgb]{0,0.5,0}{$_{\uparrow 3.9}$}&\\
\rowcolor{LightCyan}
&C$\to$M&M&\cmark&89.6&73.4&47.9&75.2&93.5&62.8&84.1&75.2\textcolor[rgb]{0,0.5,0}{$_{\uparrow 8.6}$}&\multirow{-2}{*}{73.2}\\
\midrule
AdvEnt \cite{vu2019AdvEnt}&C$\to$\{I, M\}&I&\cmark&93.3&53.0&17.2&82.8&92.2&49.3&79.6&66.8\textcolor[rgb]{1,0,0}{$_{\downarrow 0.5}$}&\multirow{2}{*}{67.0}\\
(multi-target)&C$\to$\{I, M\}&M&\cmark&87.7&65.9&29.0&73.2&91.5&47.9&75.7&67.3\textcolor[rgb]{0,0.5,0}{$_{\uparrow 0.7}$}&\\
\midrule
MTKT \cite{saporta2021multi}&C$\to$\{I, M\}&I&\cmark&93.6&54.9&18.6&84.0&\textbf{94.5}&53.4&79.2&68.3\textcolor[rgb]{0,0.5,0}{$_{\uparrow 1.0}$}&\multirow{2}{*}{69.0}\\
(multi-target)&C$\to$\{I, M\}&M&\cmark&88.3&70.4&31.6&75.9&94.4&50.9&77.0&69.8\textcolor[rgb]{0,0.5,0}{$_{\uparrow 3.2}$}&\\
\midrule
ADAS \cite{lee2022adas}&C$\to$\{I, M\}&I&\cmark&--&--&--&--&--&--&--&70.4\textcolor[rgb]{0,0.5,0}{$_{\uparrow 3.1}$}&\multirow{2}{*}{72.7}\\
(multi-target)&C$\to$\{I, M\}&M&\cmark&--&--&--&--&--&--&--&75.1\textcolor[rgb]{0,0.5,0}{$_{\uparrow 8.5}$}&\\
\midrule
\rowcolor{LightCyan}
\MTL&C$\to$\{I, M\}&I&\cmark&93.0&\textbf{60.8}&29.4&80.9&92.6&\textbf{62.3}&\textbf{85.3}&72.0\textcolor[rgb]{0,0.5,0}{$_{\uparrow 4.7}$}&\\
\rowcolor{LightCyan}
(multi-target)&C$\to$\{I, M\}&M&\cmark&90.3&75.5&\textbf{48.7}&75.3&93.6&63.2&\textbf{84.7}&75.9\textcolor[rgb]{0,0.5,0}{$_{\uparrow 9.3}$}&\multirow{-2}{*}{73.9}\\
\midrule
\rowcolor{LightCyan}
\MTA&(C$\to$M)$\to$I&I&\xmark&\textbf{95.4}&59.6&\textbf{32.7}&\textbf{86.4}&\textbf{94.5}&58.3&84.0&\textbf{72.9}\textcolor[rgb]{0,0.5,0}{$_{\uparrow 5.6}$}&\\
\rowcolor{LightCyan}
(multi-target)&(C$\to$I)$\to$M&M&\xmark&\textbf{90.5}&\textbf{75.9}&47.0&\textbf{77.9}&\textbf{95.1}&63.8&\textbf{84.7}&\textbf{76.4}\textcolor[rgb]{0,0.5,0}{$_{\uparrow 9.8}$}&\multirow{-2}{*}{\textbf{74.6}}\\
\bottomrule
\end{tabular}
}
\\\vspace{1pt}
\raggedright \scriptsize \justifying \noindent \textbf{Bold}: best IoU (\%) over all methods in each target domain.  \textcolor[rgb]{0,0.5,0}{Green} / \textcolor[rgb]{1,0,0}{red}: mIoU gain / loss \wrt the corresponding per-target baseline, marked by `*'. \Th{Extern}: using external data from the source or other target domains.
\label{tab:im}
\end{table}
%------------------------------------------------------------------------------

%------------------------------------------------------------------------------

\section{Ablation study}
\label{sec:ablation}

%------------------------------------------------------------------------------

%------------------------------------------------------------------------------
\begin{table}
\centering
\caption{Parameter analysis of $\lambda_\out$ and $\lambda_\con$ in \MTL from GTA5 (G) to CityScapes (C) and IDD (I) datasets.}
\resizebox{\linewidth}{!}{
\begin{tabular}{ccccc|ccccc}
\toprule
$\lambda_\out$&$10^{-4}$&$5\times10^{-4}$&$10^{-3}$&$10^{-2}$&$\lambda_\con$&$1$&$10$&$100$&$150$\\
\midrule
C&76.0&\textbf{76.6}&76.5&76.3&C&73.4&72.5&\textbf{76.5}&76.2\\
I&70.7&70.9&\textbf{71.2}&69.9&I&68.5&69.9&\textbf{71.2}&70.1\\
\bottomrule
\end{tabular}}
\\\vspace{1pt}
\raggedright \scriptsize \justifying \noindent 
\textbf{Bold}: best mIoU (\%) scores in each target domain.
\label{tab:parameter}
\end{table}
%------------------------------------------------------------------------------

%------------------------------------------------------------------------------
\begin{table}
\centering
\caption{Ablation study of \MTL from GTA5 (G) to CityScapes (C), IDD (I), and Mapillary (M) datasets.}
\resizebox{.9\linewidth}{!}{
\begin{tabular}{cccc|ccc|c} %
\toprule
\Th{Method}&$\mathcal{L}_\ce$&$\mathcal{L}_\out$&$\mathcal{L}_\con$&C&I&M&Avg.\\
\midrule
No adaptation&$\checkmark$&&&63.7&64.4&69.4&65.8\\
Adversarial learning&$\checkmark$&$\checkmark$&&72.8&67.5&71.9&70.7\\
Self-distillation&$\checkmark$&&$\checkmark$&75.7&69.1&74.7&73.1\\
\MTL&$\checkmark$&$\checkmark$&$\checkmark$&\textbf{76.3}&\textbf{70.8}&\textbf{75.4}&\textbf{74.1}\\
\bottomrule
\end{tabular}}
\\\vspace{1pt}
\raggedright \scriptsize \justifying \noindent  ~~~~~~~\textbf{Bold}: best mIoU (\%) scores in each target domain.
\label{tab:MTKD}
\end{table}
%------------------------------------------------------------------------------

\paragraph{\MTL}

\autoref{tab:parameter} shows how loss factors $\lambda_\out$ and $\lambda_\con$ affect performance. \MTL in general can tolerate a wide range of $\lambda_\out$ and is more sensitive to $\lambda_\con$. Based on these results, we empirically set $\lambda_\out = 10^{-3}$ and $\lambda_\con = 100$.

\autoref{tab:MTKD} shows the contribution of each component in \MTL performance. We find that adversarial learning alone cannot bring about satisfactory performance. By contrast, combining adversarial learning and self-distillation brings significant improvement.

%------------------------------------------------------------------------------
\begin{table}
\centering
\caption{Comparison of different augmentation strategies for self-distillation in \MTL from GTA5 (G) to CityScapes (C), IDD (I), and Mapillary (M) datasets.}
\resizebox{0.8\linewidth}{!}{
\begin{tabular}{c|ccc|c} %
\toprule
\Th{Method}&C&I&M&Avg.\\
\midrule
No augmentation&73.1&66.7&72.1&70.6\\
Gaussian noise w/o \MTS &73.3&66.8&72.7&70.9\\
Gaussian noise w/ \MTS &73.2&67.9&73.1&71.4\\
CutMix w/o \MTS &\textbf{76.6}&69.4&74.9&73.6\\
CutMix w/ \MTS &76.3&\textbf{70.8}&\textbf{75.4}&\textbf{74.1}\\
\bottomrule
\end{tabular}}
\\\vspace{1pt}
\raggedright \scriptsize \justifying \noindent  ~~~~~~~~~~~~~\textbf{Bold}: best mIoU (\%) scores in each target domain.
\label{tab:aug_MTKD}
\end{table}
%------------------------------------------------------------------------------

%------------------------------------------------------------------------------
\begin{figure}
\centering
\fig[.7]{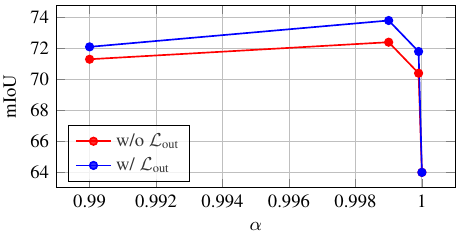}
\caption{\MTL average mIoU \vs EMA decay parameter $\alpha$ on GTA5$\to$\{CityScapes, IDD\}.}
\label{fig:alpha1}
\end{figure}
%------------------------------------------------------------------------------

%------------------------------------------------------------------------------
\begin{table}
\centering
\caption{Effect of EMA decay parameter $\alpha$ on \MTL from GTA5 (G) to CityScapes (C) and IDD (I) datasets.}
\resizebox{.9\linewidth}{!}{
\begin{tabular}{ccccccccc}
\toprule
&$\alpha$&$0$&$0.5$&$0.9$&$0.99$&$0.999$&$0.9999$&$1$\\
\midrule
&C&66.9&54.8&74.3&75.1&\textbf{75.7}&73.1&63.7\\
w/o $\mathcal{L}_\out$&I&57.2&61.1&66.8&67.5&\textbf{69.1}&67.7&64.4\\
&Avg.&62.0&57.9&70.5&71.3&\textbf{72.4}&70.4&64.0\\
\midrule
&C&34.5&32.3&60.4&75.5&\textbf{76.5}&74.2&63.7\\
w/ $\mathcal{L}_\out$&I&47.5&21.7&59.5&68.7&\textbf{71.2}&69.4&64.4\\
&Avg.&41.0&27.0&59.9&72.1&\textbf{73.8}&71.8&64.0\\
\bottomrule
\end{tabular}}
\raggedright \scriptsize \justifying \noindent  ~~~~~\textbf{Bold}: best mIoU (\%) scores in each target domain.
\label{tab:alpha1}
\end{table}
%------------------------------------------------------------------------------

\autoref{tab:aug_MTKD} shows the effect of different augmentation strategies for self-distillation in \MTL. While Gaussian noise is common~\cite{tarvainen2017mean,french2017self}, we find that CutMix is superior in cross-domain semantic segmentation. In addition, our \MTS brings further improvement by directly reducing the visual appearance shift between different domains. The combination of the two strategies brings an overall improvement of $3.5\%$ average mIoU compared with no augmentation.

\autoref{fig:alpha1} and \autoref{tab:alpha1} show how EMA decay parameter $\alpha$ affects the performance of \MTL. As the parameter $\alpha$ approaches $1$, the mIoU values increases and then drops sharply for $\alpha>0.999$. Based on these results, we empirically set $\alpha=0.999$.

%------------------------------------------------------------------------------
\begin{table}
\centering
\caption{Ablation study of \MTA from GTA5 (G) to CityScapes (C) and IDD (I) datasets.}
\resizebox{.9\linewidth}{!}{
\begin{tabular}{ccc|cc|c} %
\toprule
\Th{Method}&$\mathcal{L}_\out^\prime$&$\mathcal{L}_\con^\prime$&(G$\to$I)$\to$C&(G$\to$C)$\to$I&Avg.\\
\midrule
No adaptation&&&74.5&68.5&71.5\\
Adversarial learning&$\checkmark$&&63.9&64.8&64.3 \\
Self-distillation&&$\checkmark$&76.1&69.6&72.8 \\
\MTA&$\checkmark$&$\checkmark$&\textbf{77.0}&\textbf{70.5}&\textbf{73.7}\\
\bottomrule
\end{tabular}}
\\\vspace{1pt}
\raggedright \scriptsize \justifying \noindent  ~~~~~~~\textbf{Bold}: best mIoU (\%) scores in each target domain.
\label{tab:UTKD}
\end{table}
%------------------------------------------------------------------------------

%------------------------------------------------------------------------------
\begin{table}
\centering
\caption{Comparison of different augmentation strategies for self-distillation in \MTA from GTA5 (G) to CityScapes (C) and IDD (I) datasets.}
\resizebox{.75\linewidth}{!}{
\begin{tabular}{c|cc|c} %
\toprule
\Th{Method}&(G$\to$I)$\to$C&(G$\to$C)$\to$I&Avg.\\
\midrule
No augmentation&75.8&68.8&72.3\\
Gaussian noise &75.9&68.0&71.9\\
CutMix &\textbf{77.0}&\textbf{70.5}&\textbf{73.7}\\
\bottomrule
\end{tabular}}
\\\vspace{1pt}
\raggedright \scriptsize \justifying \noindent  ~~~~~~~~~~~~~~~~~\textbf{Bold}: best mIoU (\%) scores in each target domain.
\label{tab:aug_UTKD}
\end{table}
%------------------------------------------------------------------------------

\paragraph{\MTA}

\autoref{tab:UTKD} shows the contribution of each component in \MTA performance. An intriguing phenomenon is that adversarial learning alone is harmful. A possible explanation is that it needs the assistance of a more stable loss, as is the case of cross-entropy in \MTL. By contrast, when combined with self-distillation, it further improves performance by $0.9\%$ average mIoU, reaching a total improvement of $2.2\%$ compared with no adaptation.

\autoref{tab:aug_UTKD} shows the effect of different augmentation strategies for self-distillation in \MTA. Again, we find that CutMix works best.

As described in \autoref{sec:mta}, we train \MTA by performing self-distillation on the unseen target domain data using a consistency loss that minimizes the MSE between the student and teacher predictions, where $F'_T$ is again obtained by EMA on the parameters of $F'_S$. A simple baseline to achieve this goal is to perform knowledge distillation from a frozen teacher $F''_T$, as initialized from the pre-trained \MTL model. Accordingly, we define the \emph{frozen} consistency loss as the MSE between predictions from the student and the frozen teacher
\begin{linenomath}
\begin{align}
	\cL_\fro(X_u, F'_S) &=
		\expect_{x \sim X_u} \ell''_\con(\cA(x), F'_S)
\label{eq:con3-1}
\\
\ell''_\con(x, F'_S) &=
	\frac{1}{K} \sum_{k=1}^K \norm{F'_S(x)^{(k)} - F''_T(x)^{(k)}}^2. \label{eq:con3-2}
\end{align}
\end{linenomath}

%------------------------------------------------------------------------------
\begin{table}[!ht]
\centering
\caption{\MTA mIoU with and without frozen consistency loss from GTA5 (G) to CityScapes (C) and IDD (I) datasets.}
\resizebox{\linewidth}{!}{
\begin{tabular}{cccc|c|cc} %
\toprule
\Th{Method}&$\mathcal{L}_\out^\prime$&$\mathcal{L}_\con^\prime$&$\mathcal{L}_{\rm fro}$&(G$\to$I)$\to$C&(G$\to$C)$\to$I&Avg.\\
\midrule
No adaptation&&&&74.5&68.5&71.5\\
Adversarial&$\checkmark$&&&63.9&64.8&64.3 \\
Self-distillation&&$\checkmark$&&76.1&69.6&72.8 \\
\MTA&$\checkmark$&$\checkmark$&&\textbf{77.0}&\textbf{70.5}&\textbf{73.7}\\ \midrule
Frozen&&&$\checkmark$&73.7&66.8&70.2\\
Adversarial + Frozen&$\checkmark$&&$\checkmark$&73.9&66.8&70.3\\
Self-distillation + Frozen&&$\checkmark$&$\checkmark$&76.0&69.7&72.8\\
\MTA + Frozen&$\checkmark$&$\checkmark$&$\checkmark$&76.1&69.9&73.0\\
\bottomrule
\end{tabular}}
\\\vspace{1pt}
\raggedright \scriptsize \justifying \noindent  \textbf{Bold}: best mIoU (\%) scores in each target domain.
\label{tab:UT-KD2}
\end{table}
%------------------------------------------------------------------------------

%------------------------------------------------------------------------------
\begin{figure}
\centering
\fig[.7]{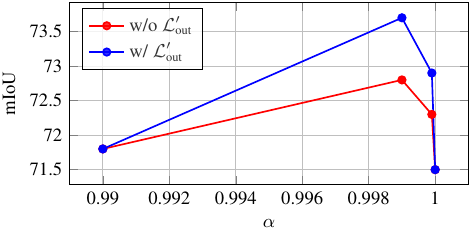}
\caption{\MTA average mIoU \vs EMA decay parameter $\alpha$ on GTA5$\to$\{CityScapes, IDD\}.}
\label{fig:alpha2}
\end{figure}
%------------------------------------------------------------------------------

\autoref{tab:UT-KD2} shows the additional ablation study of \MTA including this loss. An intriguing phenomenon is that using $\cL_\fro$~\eq{con3-1} alone is harmful, dropping performance by $1.3\%$ average mIoU compared to no adaptation. A possible explanation is that the pseudo label generated by the frozen teacher is not accurate since it is directly initialized with the pre-trained \MTL model, without refinement from EMA. Another interesting finding is that the adversarial loss $\cL'_\out$, when combined with $\cL_\fro$, is not as harmful as when used alone, which confirms its nature as an auxiliary loss. Other than that, all options involving $\cL_\fro$ are inferior to those that do not, and the best option remains $\cL'_\out + \cL'_\con$.

%------------------------------------------------------------------------------
\begin{table}
\centering
\caption{Effect of EMA decay parameter $\alpha$ on \MTA from GTA5 (G) to CityScapes (C) and IDD (I) datasets.}
\resizebox{\linewidth}{!}{
\begin{tabular}{ccccccccc}
\toprule
&$\alpha$&$0$&$0.5$&$0.9$&$0.99$&$0.999$&$0.9999$&$1$\\
\midrule
&(G$\to$I)$\to$C&3.3&37.5&68.0&75.7&\textbf{76.1}&75.4&74.5\\
w/o $\mathcal{L}_\out^\prime$&(G$\to$C)$\to$I&5.3&24.6&64.9&68.0&\textbf{69.6}&69.2&68.5\\
&Avg.&4.3&31.0&66.4&71.8&\textbf{72.8}&72.3&71.5\\
\midrule
&(G$\to$I)$\to$C&3.5&35.4&55.3&75.5&\textbf{77.0}&76.2&74.5\\
w/ $\mathcal{L}_\out^\prime$&(G$\to$C)$\to$I&5.3&34.1&59.2&68.2&\textbf{70.5}&69.7&68.5\\
&Avg.&4.4&34.7&57.2&71.8&\textbf{73.7}&72.9&71.5\\
\bottomrule
\end{tabular}}
\\\vspace{1pt}
\raggedright \scriptsize \justifying \noindent  \textbf{Bold}: best mIoU (\%) scores in each target domain.
\label{tab:alpha2}
\end{table}
%------------------------------------------------------------------------------

%------------------------------------------------------------------------------
\begin{figure}
\centering
\fig[]{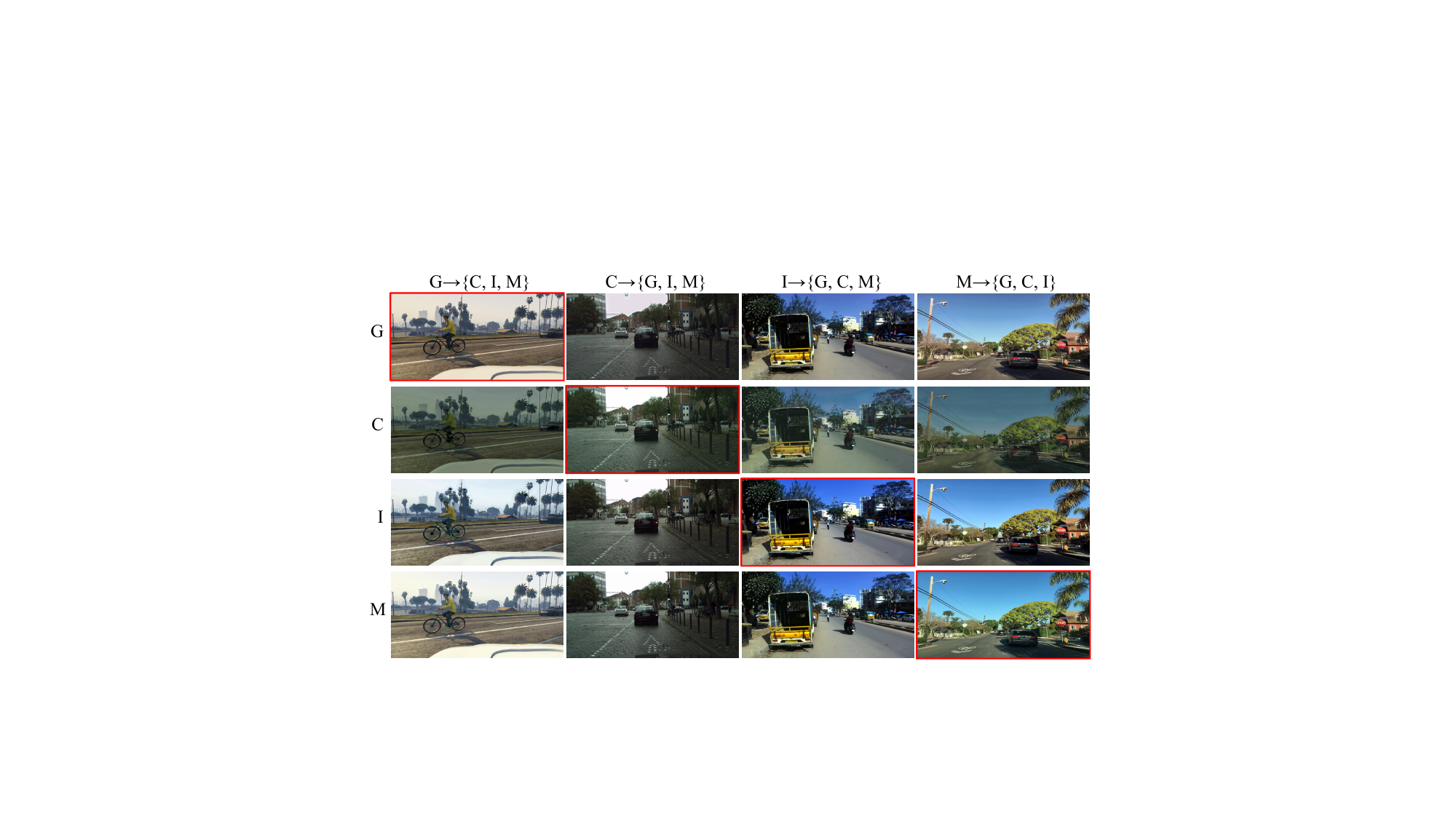}
\caption{Visual style transfer results with GTA5 (G), CityScapes (C), IDD (I), and Mapillary (M). Red-boxed images are the original inputs in each domain.}
\label{fig:trans}
\end{figure}
%------------------------------------------------------------------------------

\autoref{fig:alpha2} and \autoref{tab:alpha2} further show how EMA decay parameter $\alpha$ affects the performance of \MTA. Similar to \MTL, as the parameter $\alpha$ approaches $1$, the mIoU values increase and then drop sharply for $\alpha>0.999$. Thus, we empirically set $\alpha=0.999$.

\paragraph{\MTS}

\autoref{fig:trans} shows style transfer results between the four datasets using \MTS. We find that \MTS can learn the inherent visual style of each domain and perform synthetic-to-real, real-to-synthetic, or real-to-real style transfer between different domains.

%------------------------------------------------------------------------------

\section{Additional qualitative results}

%------------------------------------------------------------------------------
\begin{figure}[t]
\centering
\fig[.95]{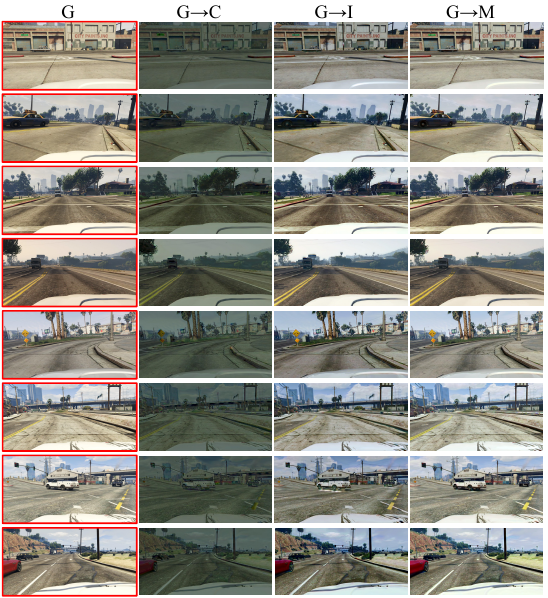}
\caption{Synthetic-to-real style transfer results from GTA5 (G) to CityScapes
(C), IDD (I), and Mapillary (M). Images in red frames are the original inputs.}
\label{fig:s2r}
\end{figure}
%------------------------------------------------------------------------------

%------------------------------------------------------------------------------
\begin{figure}[t]
\centering
\fig[.8]{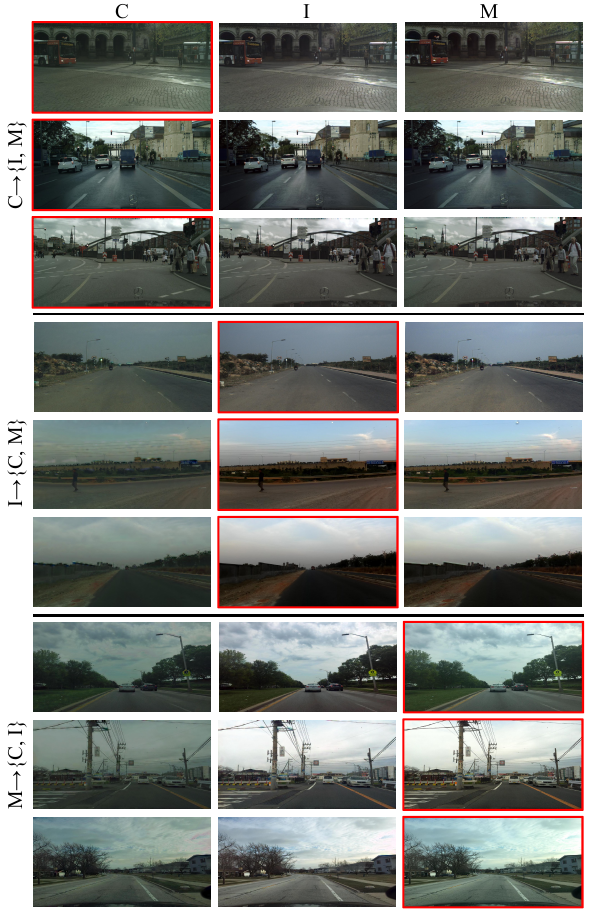}
\caption{Real-to-real style transfer results from/to CityScapes (C), IDD (I), and Mapillary (M). Images in red frames are the original inputs.}
\label{fig:r2r}
\end{figure}
%------------------------------------------------------------------------------

%------------------------------------------------------------------------------
\begin{figure*}
\centering
\fig{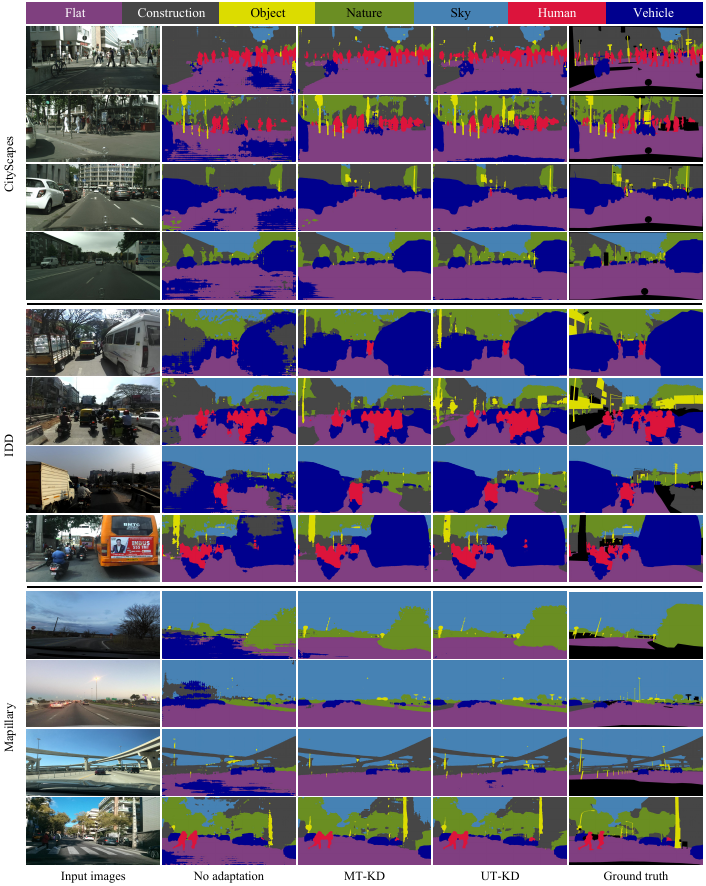}
\caption{Qualitative cross-domain semantic segmentation results from GTA5 (G) to CityScapes (C), IDD (I), and Mapillary (M) datasets. \MTL is trained on all three target domains (\ie, G$\to$\{C, I, M\}), while \MTA is initialized with the pre-trained \MTL model on two target domains and then fine-tuned on the third target domain only as unknown (\eg, (G$\to$\{C, I\})$\to$M for Mapillary).}
\label{fig:segmap2}
\end{figure*}
%------------------------------------------------------------------------------

\autoref{fig:s2r} shows more synthetic-to-real style transfer results from GTA5 to CityScapes, IDD and Mapillary using \MTS.
\autoref{fig:r2r} shows more real-to-real style transfer results from/to CityScapes, IDD, and Mapillary using \MTS. More qualitative cross-domain semantic segmentation results from GTA5 to CityScapes, IDD, and Mapillary are shown in \autoref{fig:segmap2}. \MTA can generally yield competitive or slightly better segmentation results than \MTL, although it does not access any external data. This is more evident on small objects like traffic signs and poles, as shown in the second row on the IDD dataset.

\end{document}